\documentclass[10pt,journal,compsoc]{IEEEtran}
\newif\ifpeerreview

\peerreviewfalse

\usepackage[nocompress]{cite}
\usepackage{url}
\usepackage{amsmath,amsfonts,amssymb,graphicx}
\usepackage{color}
\usepackage[inkscapelatex=false]{svg}
\usepackage{subcaption}
\usepackage{graphicx}
\usepackage{float}
\usepackage{flafter}
\usepackage{hyperref}
\graphicspath{{figures/}}

\usepackage{lipsum} 

\usepackage[switch]{lineno}

\newcommand{\paperID}{XXXX}

\title{An Attention-based Multi-Scale Feature Learning Network for Multimodal Medical Image Fusion}

\author{Meng~Zhou,
        Xiaolan~Xu,
        and~Yuxuan~Zhang
\IEEEcompsocitemizethanks{\IEEEcompsocthanksitem Meng Zhou is with the Department of Computer Science, University of Toronto, and The Hospital for Sick Children, Toronto, Canada.\protect \\
E-mail: simonzhou@cs.toronto.edu
\IEEEcompsocthanksitem Xiaolan Xu and Yuxuan Zhang are with the Department of Computer Science, University of Toronto, Canada.\protect \\
E-mail: \{landyxu,yuxuan\}@cs.toronto.edu}
}

\begin{document}

\IEEEtitleabstractindextext{%
\begin{abstract}

Medical images play an important role in clinical applications. Multimodal medical images could provide rich information about patients for physicians to diagnose. The image fusion technique is able to synthesize complementary information from multimodal images into a single image. This technique will prevent radiologists switch back and forth between different images and save lots of time in the diagnostic process. In this paper, we introduce a novel Dilated Residual Attention Network for the medical image fusion task. Our network is capable to extract multi-scale deep semantic features. Furthermore, we propose a novel fixed fusion strategy termed Softmax-based weighted strategy based on the Softmax weights and matrix nuclear norm. Extensive experiments show our proposed network and fusion strategy exceed the state-of-the-art performance compared with reference image fusion methods on four commonly used fusion metrics. Our code is available at \url{https://github.com/simonZhou86/dilran}.

\end{abstract}

}


\maketitle

\IEEEraisesectionheading{
  \section{Introduction}\label{sec:introduction}
}
%
%
%
%

\IEEEPARstart{M}{edical} imaging plays an increasingly prominent role in clinical diagnosis. Multimodal medical images can provide information from various aspects, thus helping physicians confirm the diagnosis and make decisions regarding future treatment. For example, Magnetic Resonance Imaging (MRI) images provide the soft-tissue structures of the body; computerized tomography (CT) scans provide bone structure and high-density tissue information; Positron Emission Tomography (PET), and Single-Photon Emission Computed Tomography (SPECT) images can show the metabolic activity of the cells of tissues. To acquire adequate information, physicians must analyze many different medical images, which is time-consuming and laborious. Multimodal medical image fusion (MMIF) can merge the complementary information of original images and present the required information in one fused image. This is clinically significant because physicians can now access more detailed information about disease-related changes, thus providing patients with more comprehensive and precise medical treatment and support.

In recent years, many deep learning-based approaches have been proven successful in image fusion. These methods can extract features from input images and construct a fused image with the information needed. For instance, Wang et al. \cite{wang2020multi} introduced a CNN-based medical image fusion algorithm. This method employed the trained Siamese convolutional network to fuse the pixel activity information and implemented a contrast pyramid to decompose the source images. For more generalized image fusion, Xu et al. \cite{xu2020fusiondn} proposed an unsupervised and unified densely connected network, FusionDN, which is capable of various kinds of image fusion tasks.

In this work, we propose a novel end-to-end feature learning framework for multimodal medical image fusion with clear edge information and detailed textures. To be more precise, we focus on anatomical (CT) and functional (MRI) image fusion. The framework contains a feature extractor, a fixed fusion strategy that does not involve any adjustable parameters, and image reconstruction. Our approach combines input images and generates a fused image with more detailed textures and less information loss. Our fusion results exceed the state-of-the-art performance both qualitatively in subjective vision and quantitatively in multiple fusion metrics.

The main contributions of this paper are summarized as follows:

\begin{enumerate}
    \item We propose a novel Dilated Residual Attention Network (DILRAN) for the feature extraction module. DILRAN coalesces the advantages of the residual attention network, the pyramid attention network, and the dilated convolutions. The proposed network has a faster convergence speed and can extract multi-scale deep semantic features.
    \item We introduce a novel fusion strategy, Softmax Feature Weighted Strategy, which achieves a good result and outperforms other fusion strategies.
    \item Extensive experiments show our proposed framework and fusion strategy exceed the state-of-the-art performance based on objective fusion metrics and subjective image quality.
\end{enumerate}

\section{Related Work}

There are several existing approaches for imaging fusion. Most traditional image fusions are based on the transform domain and are at pixel, feature and decision-level \cite{HERMESSI2021multimodal}. The usage of fuzzy logic and neural networks with multi-scale decomposition can handle uncertainties and improve efficiency for fusion images. Other iconic works for utilizing traditional image processing algorithms for fusion tasks include Possion Image Editing \cite{perez2003poisson}, cross bilateral filter \cite{shreyamsha2015image}, and non-subsampled contourlet transform \cite{da2006nonsubsampled}. Tian et al.\cite{tian2016multi} proposed an improved version of pulse coupled neural network (PCNN) to manage NSCT sub-images using a shallow learning approach. This new PCNN calculation determines the linking strength parameter using the local area singular value decomposition of the structural information factor. 

Recent advances in deep learning (DL) have led to the successful use of convolutional neural networks (CNN) in various imaging tasks such as classification \cite{ilyas2018noise, wang2017residual} and image super-resolution \cite{zhang2017sharp}. CNN has the ability to capture and extract features from images, which can then be used for image reconstruction. For example, Liu et al. \cite{liu2017multi} applied a CNN to multi-focus image fusion, using the network to generate a weight map of pixel activity during fusion. Hermessi et al. \cite{hermessi2018cnn} introduced a CNN-based method in the shearlet domain for extracting feature maps of high-frequency fusion. Li et al. \cite{li2019dense} proposed DenseFuse, a deep learning architecture for infrared and visible image fusion that is trained using dense blocks. Due to its limitation for only working at a single scale, Song et al.\cite{song2019msdnet} proposed a multi-scale DenseNet (MSDNet) to overcome it. They encode the multi-scale mechanism with three filters of varying sizes for effectively capturing features at different scales. Increasing the width of the encoder network can also improve the amount of detail in the fused image. Despite this, DL has great potential in the field of medical artificial intelligence due to its ability to fit complex data, its ability to learn automatically, and its multitask adaptability.

 A multi-generator multi-discriminator conditional generative adversarial network is presented by Huang et al. \cite{huang2020gan} to fuse functional information and structural information including texture details and dense structure information. This network presents better visual effects and also preserves the approximate maximum amount of information in several MMIF datasets. Fu et al. \cite{fu2021multiscale} introduced the residual pyramid attention structure: MSPRAN, which combines the advantages of residual attention and pyramid attention mechanisms in the fusion task. The framework extracts and keeps more information than a single residual attention or pyramid attention mechanism as the number of layers increases and maintains better deep features and expression capabilities. Another network structure (MSDRA) on double residual attention \cite{li2022multiscale} combines a residual network and attention to acquire important detailed features while avoiding network gradient vanish or explosion. 

\section{Proposed Method}

In this section, we provide a thorough discussion of the fusion framework we proposed and the corresponding loss function.
\subsection{Overview}

\begin{figure*}[h] 
	\begin{center}
		\includegraphics[width=.80\textwidth]{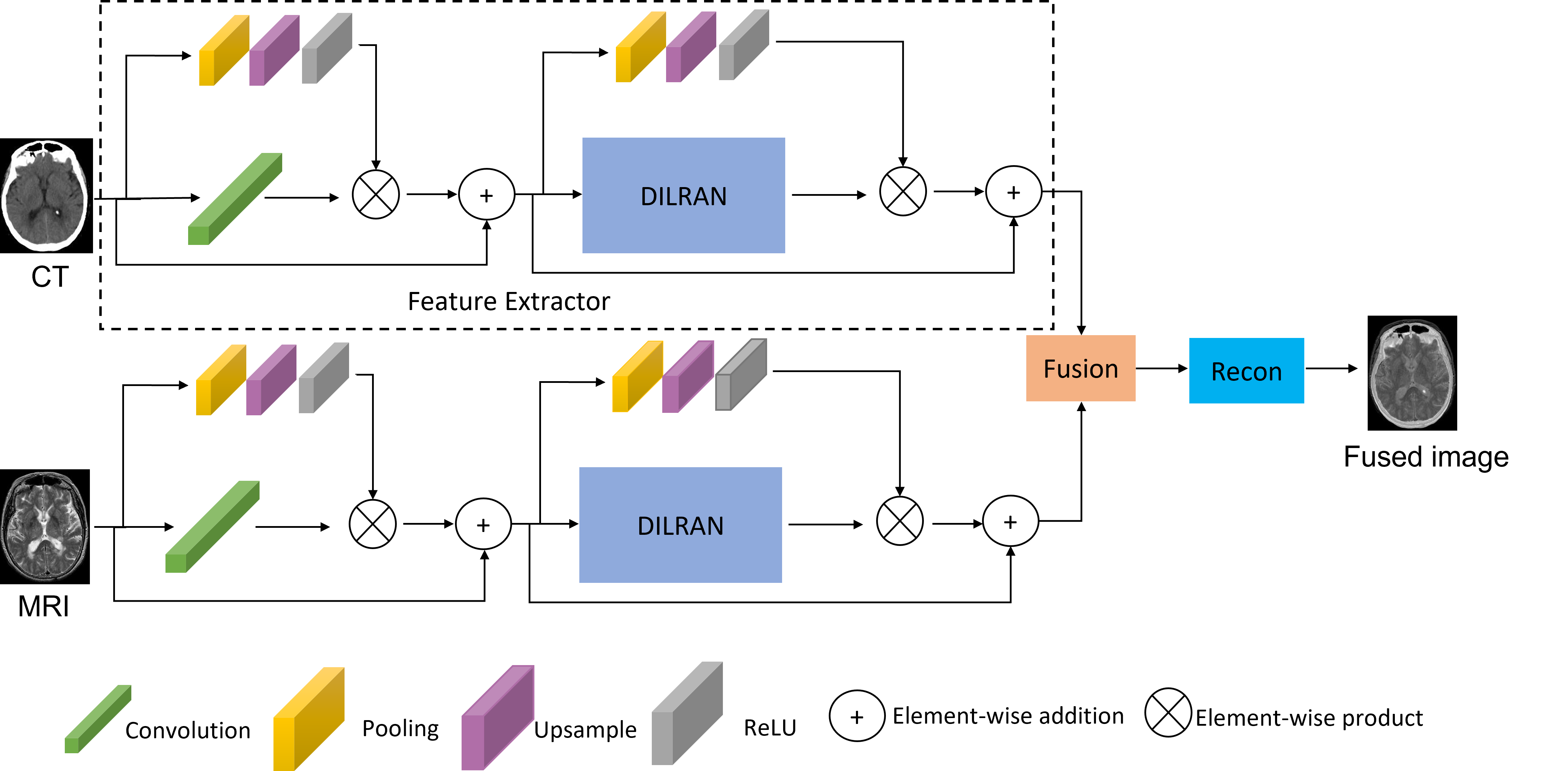}
	\end{center}
	\caption{The overall pipeline of the proposed method, which consists of a feature extractor, a fuser, and an image reconstructor}
	\label{overall}
\end{figure*}

Figure \ref{overall} summarizes the overall pipeline. We introduce a novel end-to-end framework for medical image fusion. Our proposed fusion algorithm consists of a feature extractor, a fusion module, and a reconstruction module. The feature extractor aims to formulate deep semantic features of input images ($I_1$, $I_2$), and then use them as inputs to the fusion module. We introduce Dilated Residual Attention Network (DILRAN) for the feature extraction module. Next, the fusion module aims to fuse the two extracted feature maps into one map containing features from both original feature maps. The module determines for a certain pixel in the fused feature map, whether it comes from $I_1$, $I_2$, or both. The fusion rule can have many possible solutions, i.e., $max(I_{1}^i, I_{2}^i)$ for the $i$th pixel or the weighted average of both. Finally, the reconstruction module aims to reconstruct the fused image from the output of the fusion module. A sequence of convolutional layers is used when reconstructing the final fused image. Section \ref{fe}, \ref{fs}, and \ref{recon} provide an in-depth explanation of each module we mentioned above.

\subsection{Dilated Residual Attention Network}

\begin{figure}[H] 
	\begin{center}
		\includegraphics[width=.5\textwidth]{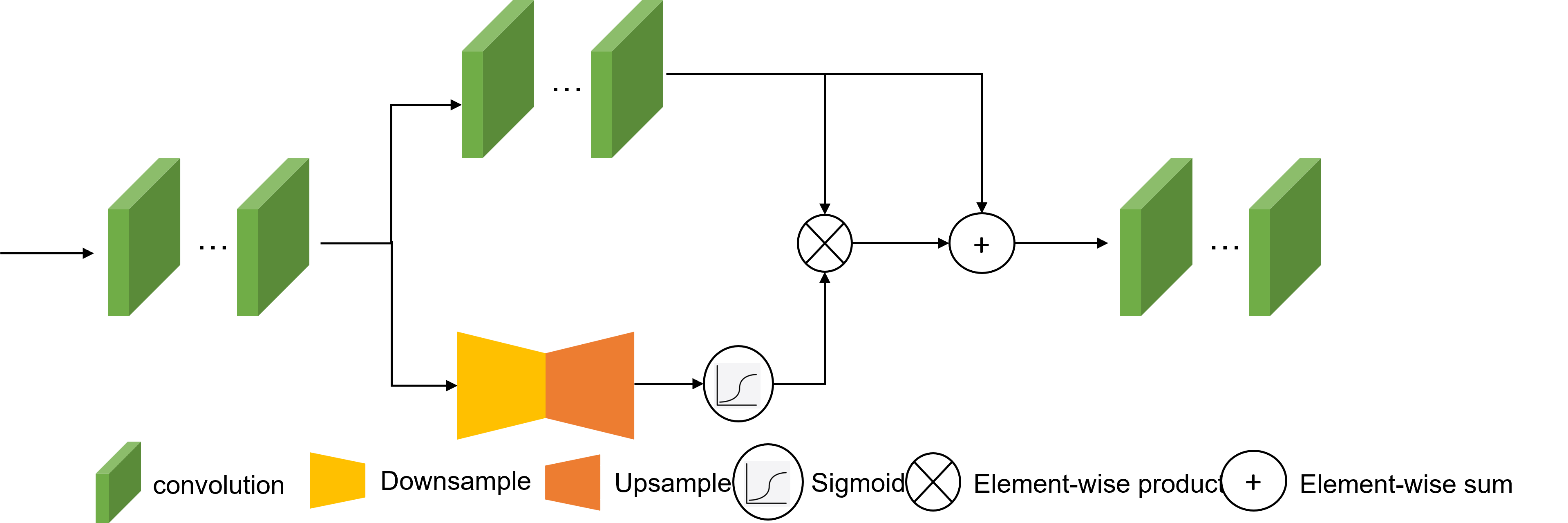}
	\end{center}
	\caption{Residual attention network architecture.}
	\label{resid_attn}
\end{figure}

The design principle of DILRAN is inspired by two state-of-the-art mechanisms: residual attention \cite{wang2017residual} and pyramid attention \cite{li2018pyramid}, as well as a recent paper \cite{fu2021multiscale} that relies on the residual pyramid attention mechanism. The classical residual attention network for image classification tasks \cite{wang2017residual}, which is realized by imposing the attention mechanism in the residual network architecture. Figure \ref{resid_attn} shows an example of the residual attention network. The top network path is called the trunk branch, which consists of a sequence of convolutional layers for feature processing. The bottom path is called the soft mask branch, which is formed by a downsample-upsample block with an activation function. The combination of the trunk and the soft mask branch build up the residual attention network following Equation \eqref{resid_eq}.

\begin{equation} \label{resid_eq}
    F(x) = (1 + S(x)) * T(x)
\end{equation}

Where $x$ is the feature map from the previous layer, $F(\cdot)$ is the output feature map, $S(\cdot)$ is the function for the soft mask branch, and $T(\cdot)$ is the convolution operation.

\begin{figure}[h] 
	\begin{center}
		\includegraphics[width=.5\textwidth]{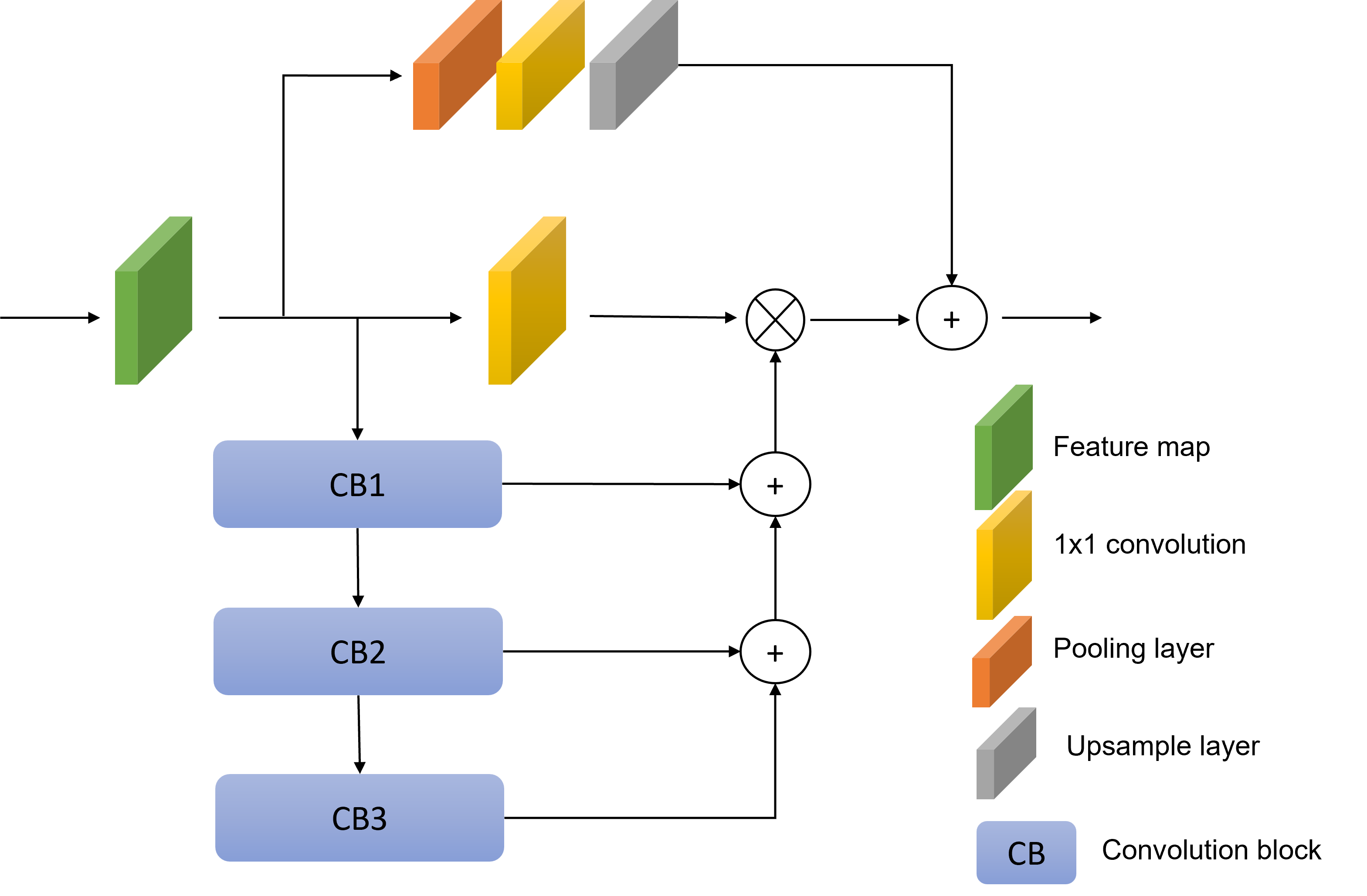}
	\end{center}
	\caption{Pyramid attention network architecture.}
	\label{pyram_attn}
\end{figure}

The advantage of the residual attention network is that it has a faster convergence speed without gradient vanishing or explosion, however, it may not be able to extract and learn deep features. Hence, the other mechanism we adopt is the pyramid attention network \cite{li2018pyramid}. To be more specific, we focus on the feature pyramid attention that could provide deeper semantic features and attention advantages to the output feature map. A sample pyramid attention network architecture is shown in Figure \ref{pyram_attn}. The convolution block usually contains a sequence of convolutional layers that capture features at different scales and receptive fields. We adapt the idea to replace convolutions with larger kernel filters with a sequence of convolutions with smaller kernel filters \cite{szegedy2016rethinking}. Hence, the $CB1$ in our method is a single $3 \times 3$ convolutions; the $CB2$ consists of \textit{two} $3 \times 3$ convolutions (represents a single $5 \times 5$ convolutions), and the $CB3$ consists of \textit{three} $3 \times 3$ convolutions (represents a single $7 \times 7$ convolutions). The output features can be computed by the following Equation \eqref{pyr_attn}.

\begin{equation} \label{pyr_attn}
    F(x) = (1 + P_1(P_2(P_3(x)))) * C(x)
\end{equation}

Where $x$ is the feature map from the previous layer, $F(\cdot)$ is the output feature map, $P_i(\cdot), i \in [1,2,3]$ is the parameters of the corresponding $CB_i, i\in [1,2,3]$ in the feature pyramid attention network, $C(\cdot)$ is the convolution operation.

\begin{figure}[h] 
	\begin{center}
		\includegraphics[width=.5\textwidth]{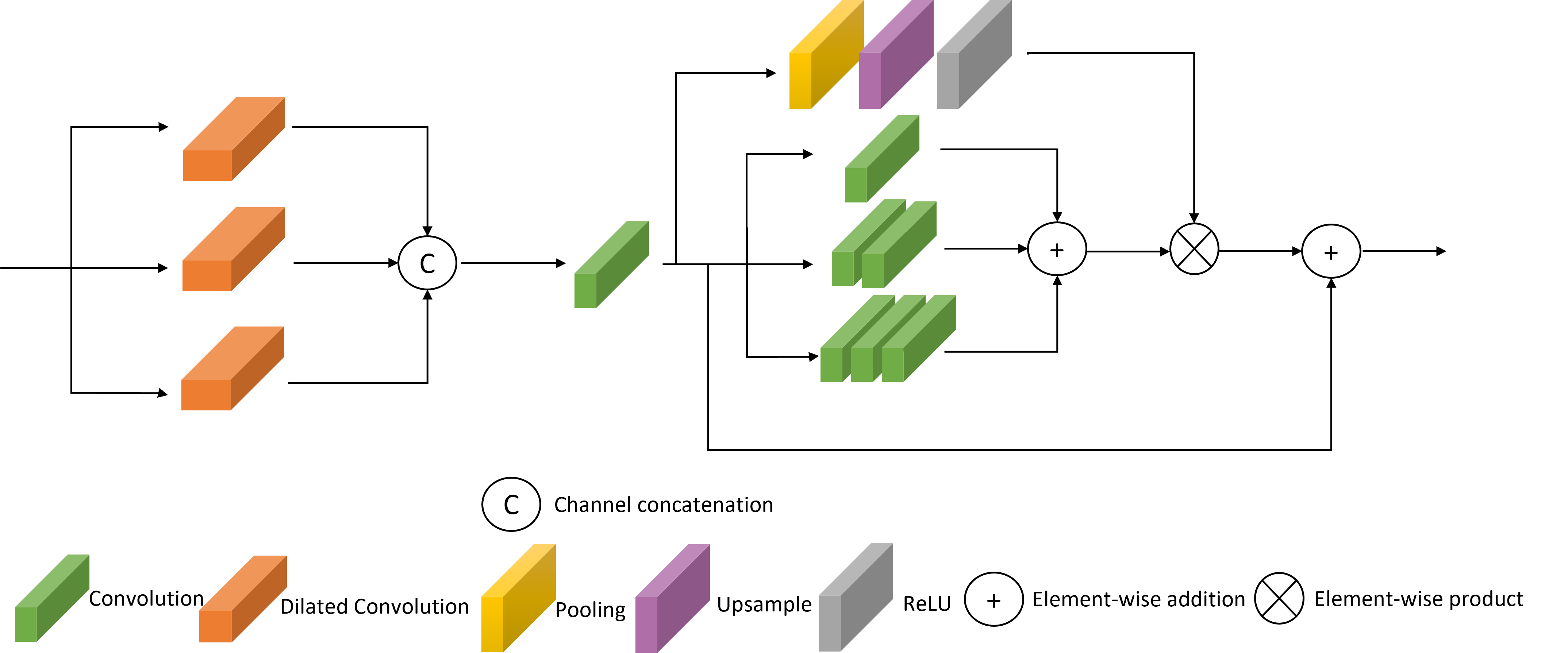}
	\end{center}
	\caption{Proposed Dilated Residual Attention Network architecture. Three pyramid convolution blocks contain one, two, and three consecutive $3 \times 3$ convolutional layers, respectively.}
	\label{dilran}
\end{figure}

In \cite{fu2021multiscale}, they sequentially downsample the input feature maps to formulate the pyramid structure, and upsample again when performing the element-wise summation as shown in Figure \ref{pyram_attn}. However, this may lose the local information and fine details in the image. To solve this problem, we leverage the ${\{1,3,5\}}$-dilated convolution \cite{yu2015multi} on shallow features of the original input image to extract the multi-scale information instead of downsampling the feature map. The receptive field is expanded using three different dilated convolutions to improve the discriminative multi-scale feature extraction ability of the model. Once the multi-scale features are extracted, we concatenate those features channel-wise, and then the residual-pyramid attention paradigm is used to further extract deep features. The deep features are the output of the feature extraction module and are used in both the fusion and reconstruction modules. Our designed DILRAN architecture is shown in Figure \ref{dilran}.

\subsection{Feature Extraction Module} \label{fe}

The detailed architecture for the feature extractor is shown in Figure \ref{overall}. The input image is first passed to a residual attention network with a $1 \times 1$ convolutional layer to obtain a 64-dimensional shallow feature map. Then, the shallow feature is passed to the DILRAN module to obtain deep semantic features. To ensure the final output feature map contains both the local and global information of the original input image, and to stabilize the training process, another residual attention network is utilized between shallow and deep semantic features. Finally, the output feature map is obtained and used in the following modules.

\subsection{Fusion Strategy} \label{fs}
The fusion strategy in the fusion module is used to fuse the extracted features of input images into a single feature map. In this section, We introduce a novel fusion strategy termed ``Softmax Feature Weighted Strategy'' that exceeds the state-of-the-art performance. The advantage of the proposed fusion strategy is validated in Section \ref{exp_res}. 

\subsubsection{Softmax-based weighted strategy}
We obtain two output feature maps $f_1, f_2$ from the extraction module for input images $I_1, I_2$, respectively. The output feature map from the extraction module can be used to generate the corresponding weight map that indicates the amount of contribution of each pixel to the fused feature map \cite{lahoud2019zero}. First, to get the weight map, we take the Softmax \cite{bridle1989training} operation to the feature map which can be realized by Equation \eqref{softmax_op}, where $x_i$ is the $i$th channel of the output feature map $x$.

\begin{equation} \label{softmax_op}
    S(x_i) = \frac{\exp(x_i)}{\sum_{i} \exp(x_i)}
\end{equation}

After we obtained the Softmax output, we compute the matrix nuclear norm ($\|\cdot\|_{*}$), which is the summation of its singular values. Finally, we obtain the weights for the output feature map by taking the weighted average of the maximum value of the nuclear norm. The formula is given in Equation \eqref{sfnn}.

\begin{align} \label{sfnn}
    &W_1 = \frac{\phi(\|S(x_{i})^1\|_{*})}{\phi(\|S(x_{i})^1\|_{*}) + \phi(\|S(x_{i})^2\|_{*})} \\
    &W_2 = \frac{\phi(\|S(x_{i})^2\|_{*})}{\phi(\|S(x_{i})^1\|_{*}) + \phi(\|S(x_{i})^2\|_{*})}
\end{align}

Where $S(x_{i})^j, j \in [1,2]$ is the weight map after Softmax operation for the feature map $f_j, j \in [1,2]$, $\phi(\cdot)$ can be any arbitrary functions. The final fused feature map is then given by $f = W_1 * f_1 + W_2 * f_2$.

\subsection{Reconstruction Module} \label{recon}

\begin{figure}[h] 
	\begin{center}
		\includegraphics[width=.5\textwidth]{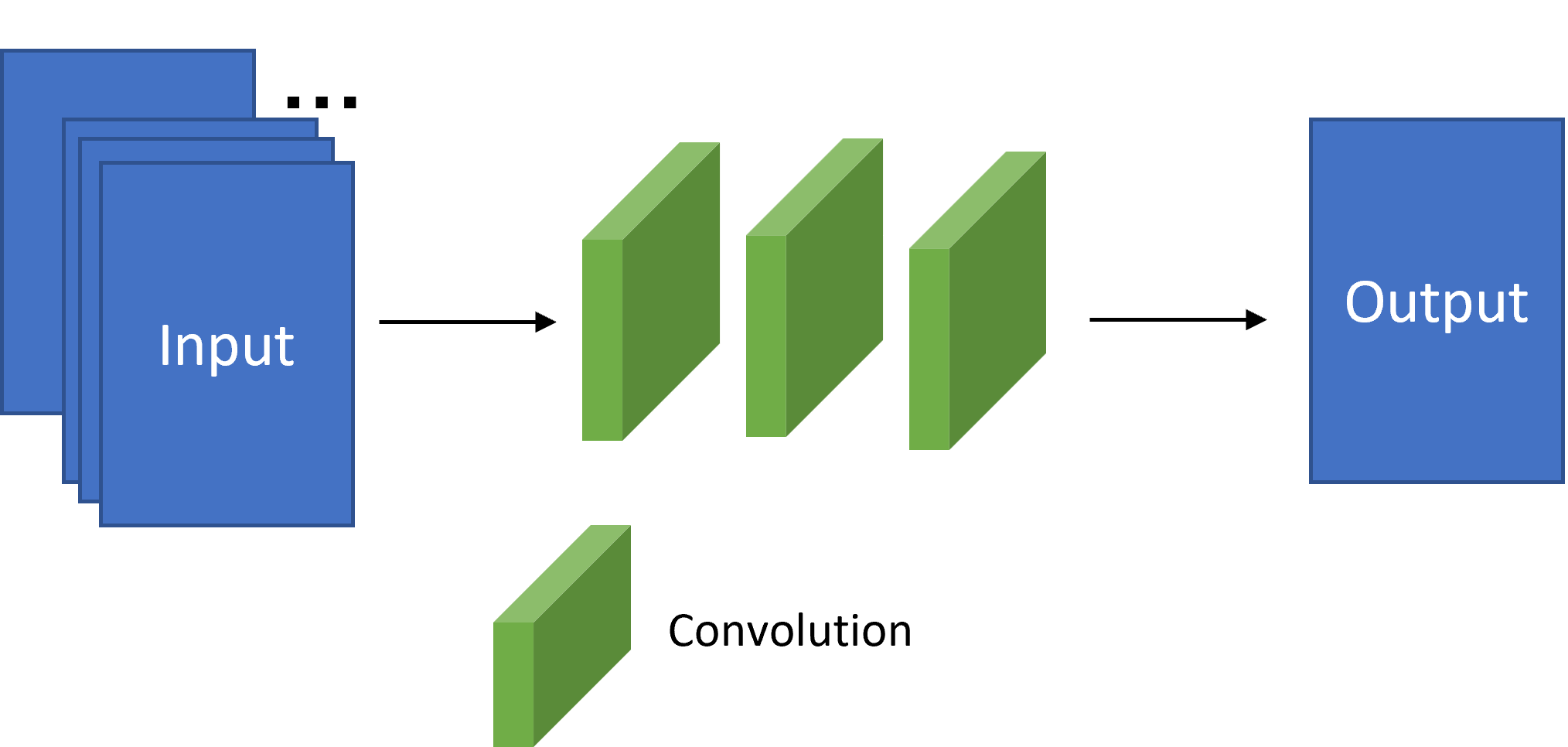}
	\end{center}
	\caption{Image reconstruction module architecture. There are three consecutive $3 \times 3$ convolutional layers, each with the output channel 64, 32, and 1, respectively.}
	\label{img_recon}
\end{figure}

The input of the reconstruction module is the fused feature map from the fusion module. The reconstruction module is used to generate the fused human-visible image from the fused feature map while retaining as many details as possible. Our reconstruction module utilizes three consecutive $3 \times 3$ convolutional layers with output channels 64, 32, and 1 to reduce the channel from 64 to 1, indicating the output image is gray-scale. An overview of the reconstruction module is shown in Figure \ref{img_recon}.

\subsection{Loss function} \label{loss_func}

We hypothesize that in order to make the fused image as close as two input images, we need to minimize the distance between the image reconstructed from deep semantic features and the original input image. Then, after we apply the fusion strategy and obtain the fused deep features, the reconstructor will produce a more realistic and reasonable fused image. Our goal is to coalesce input images to form an output fused image that retains as many details as the input. In other words, the fused image should have a strong correlation with both input images. We argue that the gap between fused images and input images can be mainly measured by the common regression loss function ``mean squared error'' (MSE). We also add gradient loss \cite{ma2020structure, fabbri2018enhancing} to model the fine details of textures in the reconstructed image and perceptual loss \cite{johnson2016perceptual} to model the high-level semantic similarity between reconstructed and input images. In detail, our loss function is defined as follows:


\begin{equation} \label{mse}
    MSE = \sum_{j=1}^{M} \|I_o - I_j\|_{F}^{2}
\end{equation}

\begin{equation} \label{grad_img}
    \nabla_I = \sum_{j=1}^{M} \|\nabla I_o - \nabla I_j\|_{2}^{2}
\end{equation}

\begin{equation} \label{perp_loss}
    Percep = \frac{1}{C*W_i*H_i} \sum_{j=1}^{M} \sum_{k=1}^{C} \|f_{i}^{k}(I_o) - f_{i}^{k}(I_j)\|_{2}^{2}
\end{equation}

\begin{equation} \label{total_loss}
    \mathcal{L}(\theta) = \frac{1}{W*H} (MSE + \lambda_1 * \nabla_I) + \lambda_2 * Percep 
\end{equation}

Equation \eqref{mse} is a variant of the MSE loss, where $M$ is the number of input images, $I_o$ is the output image, $\|\cdot\|_F$ is the matrix Frobenius norm. Image gradient loss is given in Equation \eqref{grad_img}, which is realized by the $\mathcal{L}_2$ norm of the image gradient in $x$ and $y$-direction. Equation \eqref{perp_loss} is the perceptual loss \cite{johnson2016perceptual}, where $f_{i}^k(x)$ is the the $k$th channel in $i$th layer (with size $W_i \times H_i$) from the pre-trained VGG16 network \cite{simonyan2014very} with input image $x$, and $C$ is the number of channels. We prefer $i$ to be large, i.e., the deeper layer of the VGG network. Finally, the total loss function is given in Equation \eqref{total_loss}, $\theta$ is the set of network weights to be optimized, $\lambda_1, \lambda_2$ are weight balancing factors of the gradient and perceptual loss, respectively. The total loss has been normalized by the total pixels in the image with width $W$ and height $H$.

\section{Experimental Results} \label{exp_res}

In this section, we will discuss the dataset we used, the experimental setup, evaluation metrics, and the experimental results we obtained.
\subsection{Data}

We adapt the commonly used medical image fusion dataset ``The Harvard Whole Brain Atlas'' \footnote{Available at \url{https://www.med.harvard.edu/aanlib/}} \cite{summers2003harvard} for this work. The dataset we obtained contains three sets of co-registered multimodal medical image pairs: MRI-CT (184 pairs), MRI-PET (269 pairs), and MRI-SPECT (357 pairs). All images are with size $256 \times 256$, MRI and CT are single-channel images with pixel intensities in the range of $[0, 255]$, and PET and SPECT are three-channel images with pixel intensities in the range of $[0, 255]$. In terms of the computational complexity of our model and the time given for this work, we only focus on the MRI-CT fusion task.

\subsection{Experimental Setup}

For the MRI-CT fusion task, we have 184 image pairs. We randomly select 20 image pairs for testing, so that these data are not included in the training process. For the rest of the data, we further split to training and validation data with the ratio of $0.8:0.2$, i.e., 80\% of the data for training, 20\% for validation. All images are normalized in the range of $[0, 1]$ before training. During the training phase, we take out the fusion module, and only train the feature extractor and reconstructor as discussed in Section \ref{loss_func}. The stochastic gradient descent with Adam update rule \cite{kingma2014adam} is utilized as the optimizer, the learning rate is set to 0.0001 and the model is trained for 100 epochs. The weight balancing factors $\lambda_1$ and $\lambda_2$ for image gradient loss and perceptual loss are all set to 0.2, and we use the third VGG block \cite{simonyan2014very} output to calculate the perceptual loss. To prove the effectiveness of the proposed method, we compare it with the following methods: zero-shot learning for medical image fusion \cite{lahoud2019zero}, MSRPAN \cite{fu2021multiscale}, and MSDRA \cite{li2022multiscale}. The parameters of these methods are the default setting suggested by the authors. The experimental environment is a Windows 10 (64 bit) system computer with the Intel Core I7 CPU, NVIDIA GeForce RTX 3050 Ti GPU, and 16 GB of RAM. 

\subsection{Evaluation Metrics}

There are many fusion metrics have been proposed in the past few years. However, different metrics reflect the different perspectives of fusion performance. Hence, we select six different, but commonly used fusion metrics in this work: Peak signal-to-noise ratio (PSNR), Structural Similarity (SSIM) \cite{wang2004image}, Feature SSIM (FSIM) \cite{zhang2011fsim}, Mutual information (MI) \cite{qu2002information}, pixel-wise feature MI ($FMI_{pixel}$) \cite{haghighat2011non}, and Information Entropy (EN). PSNR is the ratio between the maximum signal power and the signal noise power. It is used to measure the quality of the reconstructed image. The larger the PSNR value is, the better the output image quality. Since PSNR is a reference-based metric, for input image $I_1, I_2$ and fused image $I_o$, we first compute the $PSNR(I_1, I_o)$ and $PSNR(I_2, I_o)$, and then the average of these two numbers to obtain the final PSNR value. SSIM is a metric to measure the structural similarity of two images from brightness, contrast, and structure. The SSIM is calculated in the same way as the PSNR. FSIM is a full reference fusion metric that measures how close the phase congruency and magnitude of the gradient are between the fused image and source image(s). MI measures the similarity of the image intensity distributions between two or more images, which is calculated in the same way like the PSNR as well. Pixel-wise feature MI is a non-reference image fusion metric that measures the mutual information of the image features between fused image and source image(s). Finally, information entropy measures the amount of information in the fused image. For all metrics, the larger the value, the better the fusion quality.

\subsection{Selection of fusion strategy}

\begin{figure}[h!]
     \centering
          \begin{subfigure}[b]{0.24\textwidth}
         \centering
         \includegraphics[width=\textwidth]{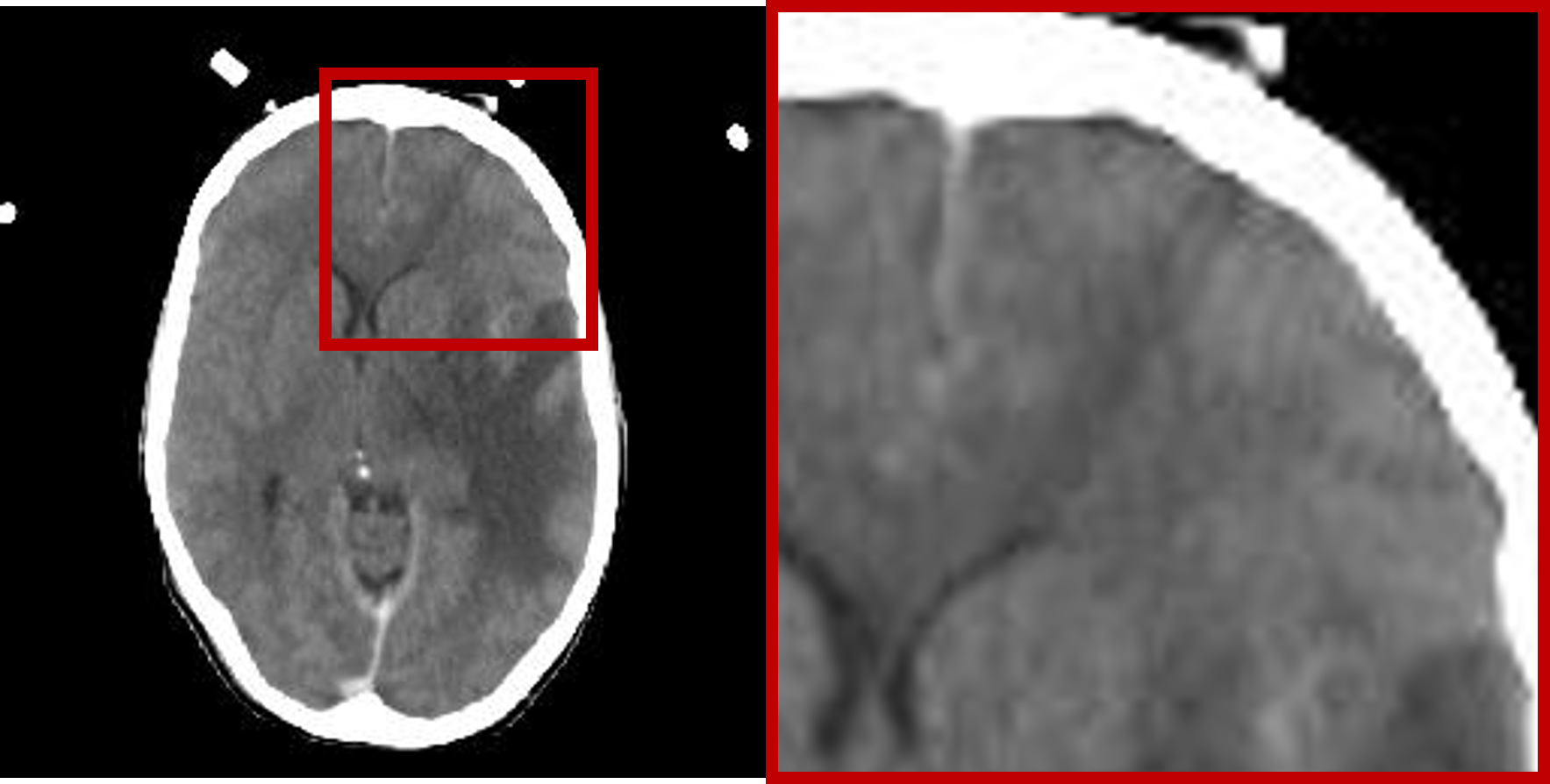}
         \caption{CT}
         \label{fig:ct_orig}
     \end{subfigure}
     \hfill
          \begin{subfigure}[b]{0.24\textwidth}
         \centering
         \includegraphics[width=\textwidth]{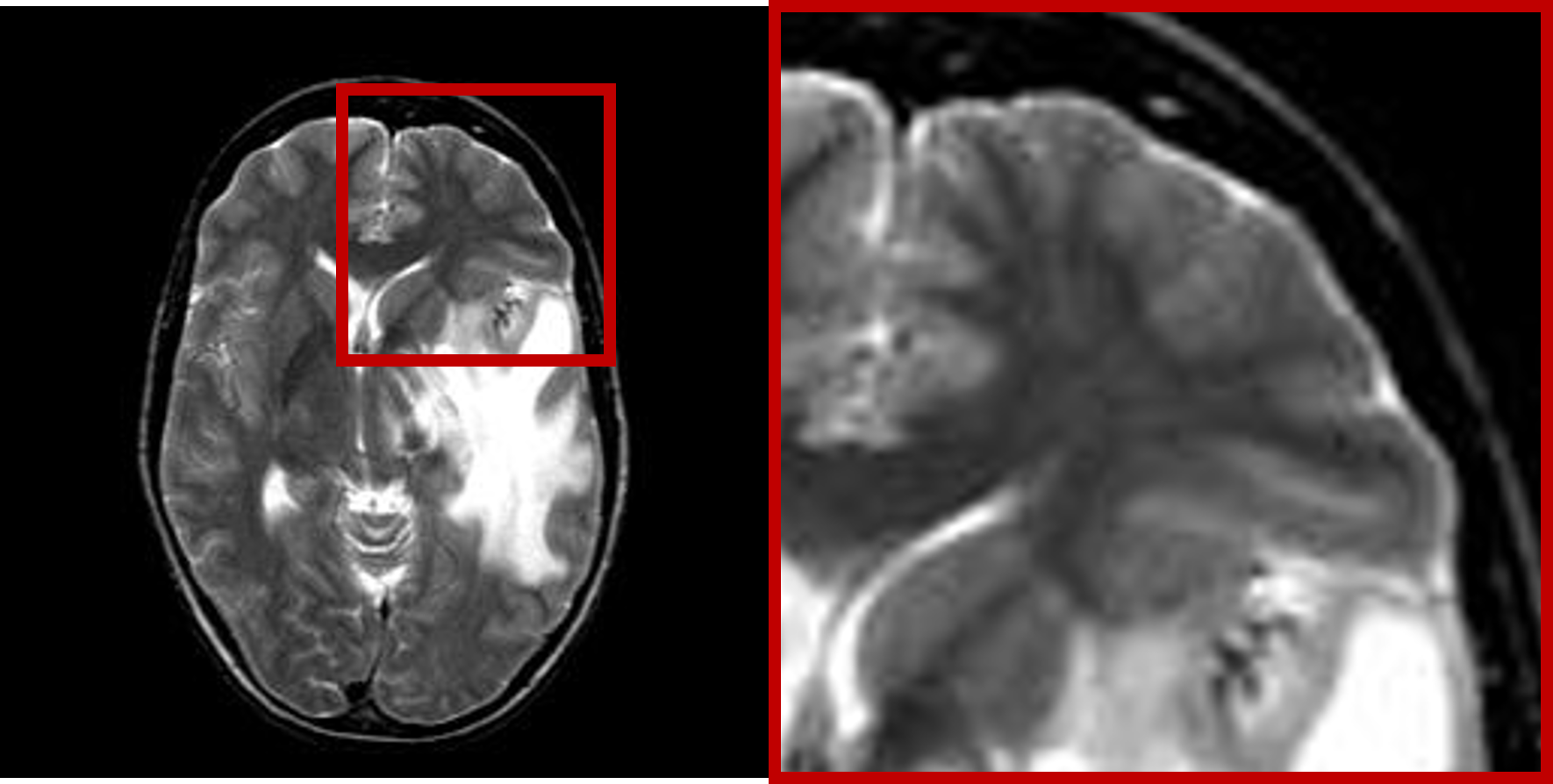}
         \caption{MRI}
         \label{fig:mri_orig}
     \end{subfigure}
     \hfill
     \begin{subfigure}[b]{0.24\textwidth}
         \centering
         \includegraphics[width=\textwidth]{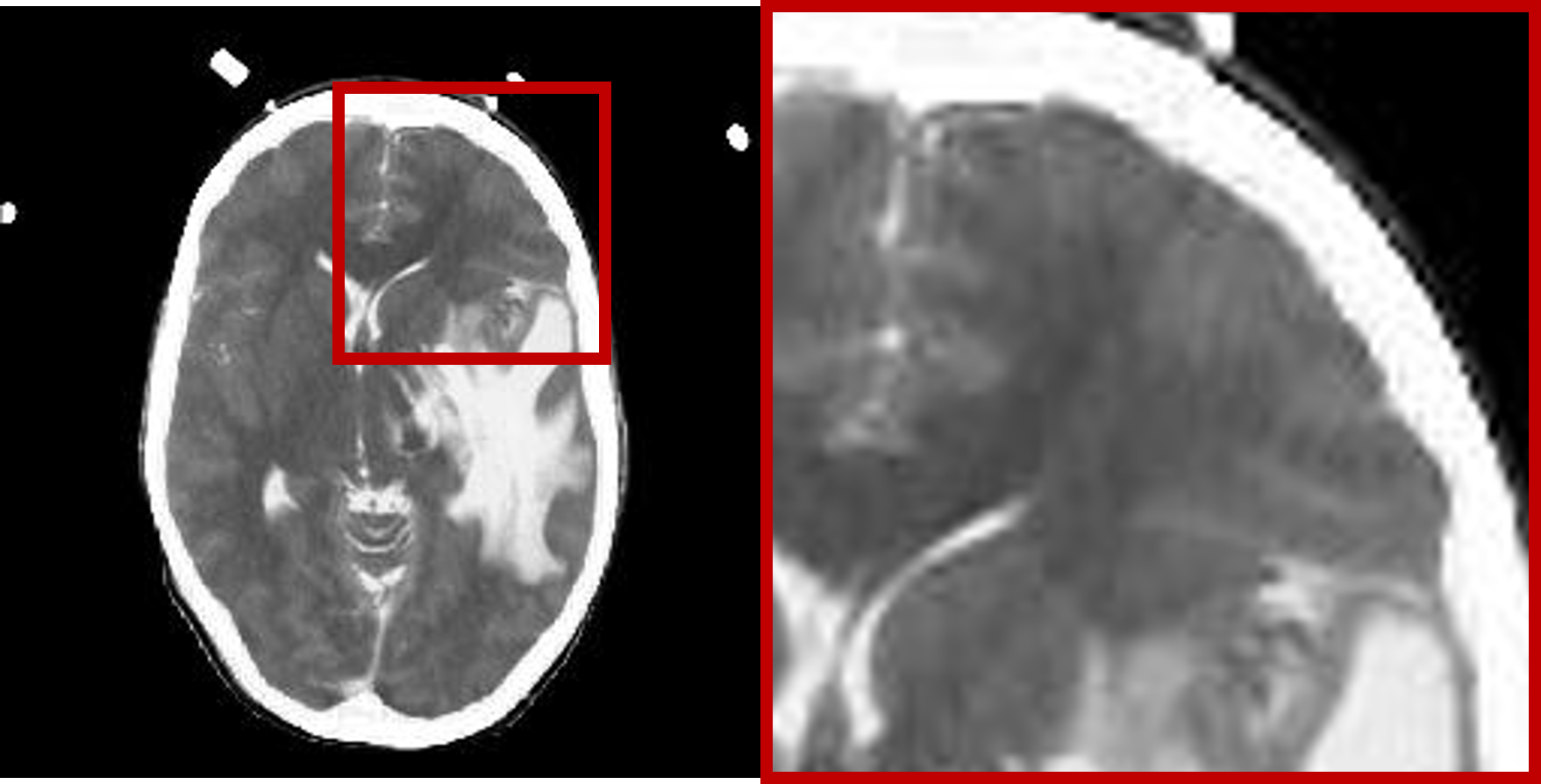}
         \caption{FER \cite{fu2021multiscale}}
         \label{fig:fer}
     \end{subfigure}
     \hfill
     \begin{subfigure}[b]{0.24\textwidth}
         \centering
         \includegraphics[width=\textwidth]{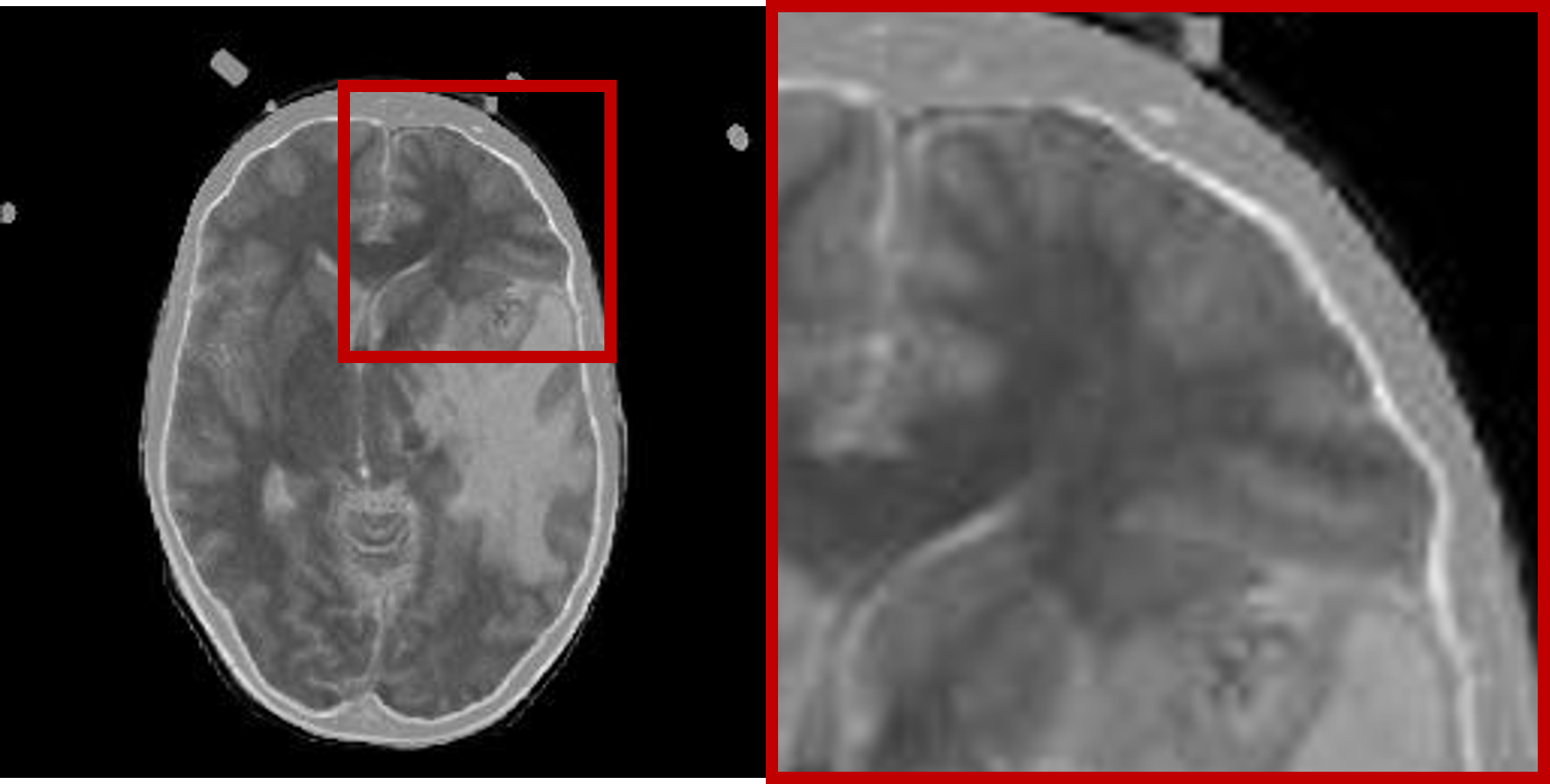}
         \caption{FL1N \cite{li2022multiscale}}
         \label{fig:fl1n}
     \end{subfigure}
     \hfill
     \begin{subfigure}[b]{0.24\textwidth}
         \centering
         \includegraphics[width=\textwidth]{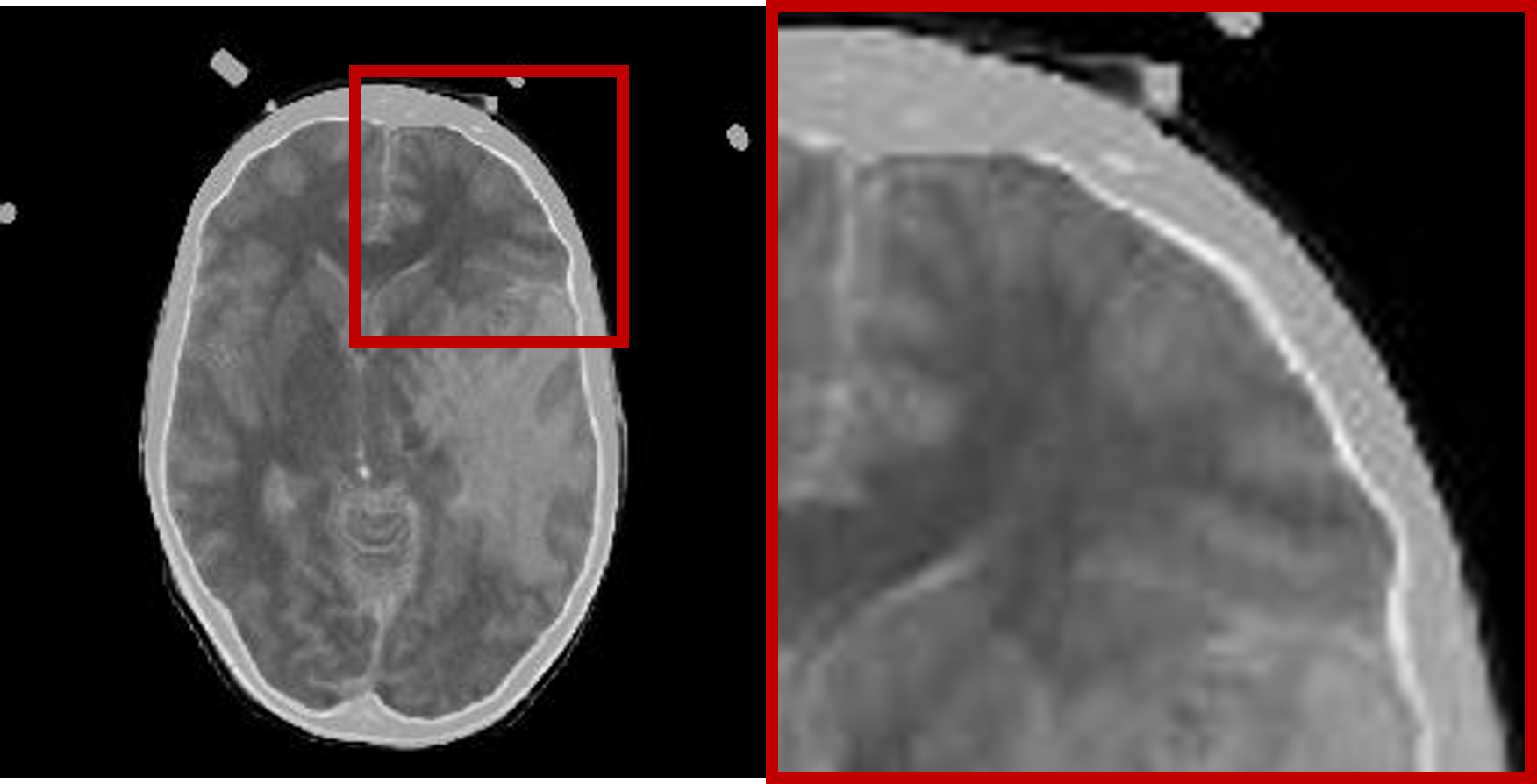}
         \caption{SFNN-Max}
         \label{fig:sfnnMa}
     \end{subfigure}
     \hfill
     \begin{subfigure}[b]{0.24\textwidth}
         \centering
         \includegraphics[width=\textwidth]{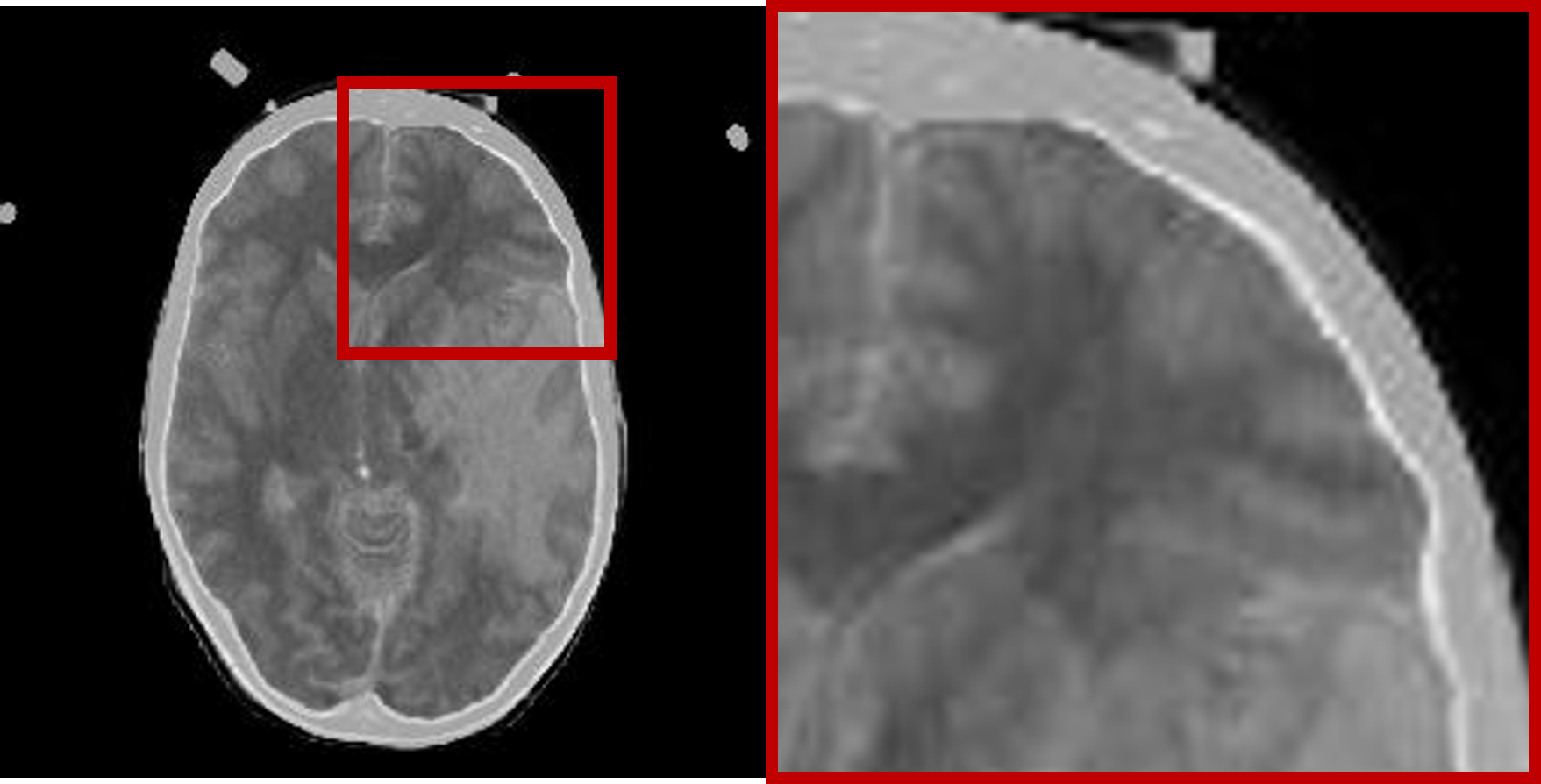}
         \caption{SFNN-Mean}
         \label{fig:sfnnMe}
     \end{subfigure}
     \caption{Qualitative comparison of different fusion strategies between CT and MRI. (a) and (b) are source images, (c) is the FER fusion strategy \cite{fu2021multiscale}, (d) is the FL1N fusion strategy \cite{li2022multiscale}, and (e), (f) are the strategies we proposed in this work.}
     \label{fs_qualres}
\end{figure}

As we discussed in Section \ref{fs} and Equation \eqref{sfnn}, our candidate functions for $\phi(\cdot)$ are $max()$, $mean()$, and $sum()$ functions to determine the weights from the matrix nuclear norm. We also investigate the squared version of the three functions above, e.g., $\phi(\cdot) = max()^2$. Next, we provide detailed quantitative results when different fusion strategies are used in Table \ref{dif_fs}. FER \cite{fu2021multiscale} and FL1N \cite{li2022multiscale} are two fusion strategies proposed previously for the same task. The SFNN is what we proposed in this work, and selection methods are described above. Our proposed fusion strategy performed well on four metrics (PSNR, FMI, FSIM, EN). Figure \ref{fs_qualres} shows the qualitative results of different fusion strategies. For SFNN, we select two strategies that produce the best image quality visually, and also notice that different choices of $\phi$ in our proposed strategy do not affect the metrics very much. Although our SFNN-$max^2$ achieves the best result in terms of the objective metrics, it has a low visual fidelity, hence we select the second optimal SFNN-$max$ as our ultimate strategy in this work. Compared with the FER strategy \cite{fu2021multiscale} in Figure \ref{fig:fer}, our results have better fidelity, and the inner tissue boundary is more clear. Similarly, our results produce a more brightness edge than FL1N strategy \cite{li2022multiscale} in Figure \ref{fig:fl1n}, which is better when separating between the edge and tissues.

\begin{table}[H]
\renewcommand{\arraystretch}{1.3}
\caption{Comparison between different fusion strategies, bold values represent the best results, and values in blue represent the second-best results.}
\centering
\resizebox{\columnwidth}{!}{
\begin{tabular}{ccccccc}
\hline & PSNR & SSIM & MI & FMI-Pixel & FSIM & Entropy \\
\hline FER \cite{fu2021multiscale} & $13.944$ & $0.736$ & $4.551$ & $0.876$ & $0.806$ & $8.720$ \\
FL1N \cite{li2022multiscale} & $15.979$ & $0.739$ & $4.569$ & $0.878$ & $0.813$ & $9.782$ \\
SFNN ($mean$) & $15.876$ & ${\color{blue}0.740}$ & $\textbf{4.578}$ & $0.876$ & $0.812$ & $9.772$ \\  
SFNN ($mean^2$) & $16.311$ & $\textbf{0.741}$ & ${\color{blue}4.567}$ & ${\color{blue}0.883}$ & $0.819$ & $9.812$ \\
SFNN ($max$) & ${\color{blue}16.413}$ & ${\color{blue}0.740}$ & $4.558$ & $\textbf{0.891}$ & ${\color{blue}0.820}$ & ${\color{blue}9.816}$ \\
SFNN ($max^2$) & $\textbf{18.258}$ & ${\color{blue}0.740}$ & $4.451$ & $\textbf{0.891}$ & $\textbf{0.829}$ & $\textbf{9.887}$ \\
SFNN ($sum$) & $15.876$ & ${\color{blue}0.740}$ & $\textbf{4.578}$ & $0.876$ & $0.812$ & $9.772$ \\
SFNN ($sum^2$) & $16.311$ & ${\color{blue}0.740}$ & $\textbf{4.578}$ & $0.876$ & $0.819$ & $9.812$ \\
\hline
\end{tabular}
}
\label{dif_fs}
\end{table}

\subsection{Comparison with other methods}

Table \ref{main_res} presents the quantitative results of the MRI-CT fusion task between our proposed method and other baseline methods. Our method performs well on multiple fusion metrics: PSNR, pixel-wise FMI, FSIM, and Entropy. The largest PSNR value indicates our fused image contains less noise and achieves the best image quality. The largest FMI value demonstrates our fused image contains as many source image features as possible. The largest FSIM value suggests our fused image has less information loss at the feature level. Finally, the largest entropy shows our fused image contains more information and details. 

\begin{table}[h]
\renewcommand{\arraystretch}{1.3}
\caption{Comparison between different methods, bold numbers represent optimal values}
\centering
\resizebox{\columnwidth}{!}{
\begin{tabular}{ccccccc}
\hline & PSNR & SSIM & MI & FMI-Pixel & FSIM & Entropy \\
\hline Zero-shot \cite{lahoud2019zero} & $13.525$ & $0.681$ & $4.633$ & $0.836$ & $0.738$ & $4.279$ \\
MSDRA \cite{li2022multiscale} & $15.693$ & $0.697$ & $4.586$ & $0.867$ & $0.797$ & $8.167$ \\
MSRPAN \cite{fu2021multiscale} & $14.528$ & $\textbf{0.741}$ & $\textbf{4.652}$ & $0.874$ & $0.808$ & $8.969$ \\
Ours & $\textbf{16.413}$ & $0.740$ & $4.558$ & $\textbf{0.891}$ & $\textbf{0.820}$ & $\textbf{9.816}$ \\
\hline
\end{tabular}
}
\label{main_res}
\end{table}

\begin{figure}[h!]
     \centering
          \begin{subfigure}[b]{0.24\textwidth}
         \centering
         \includegraphics[width=\textwidth]{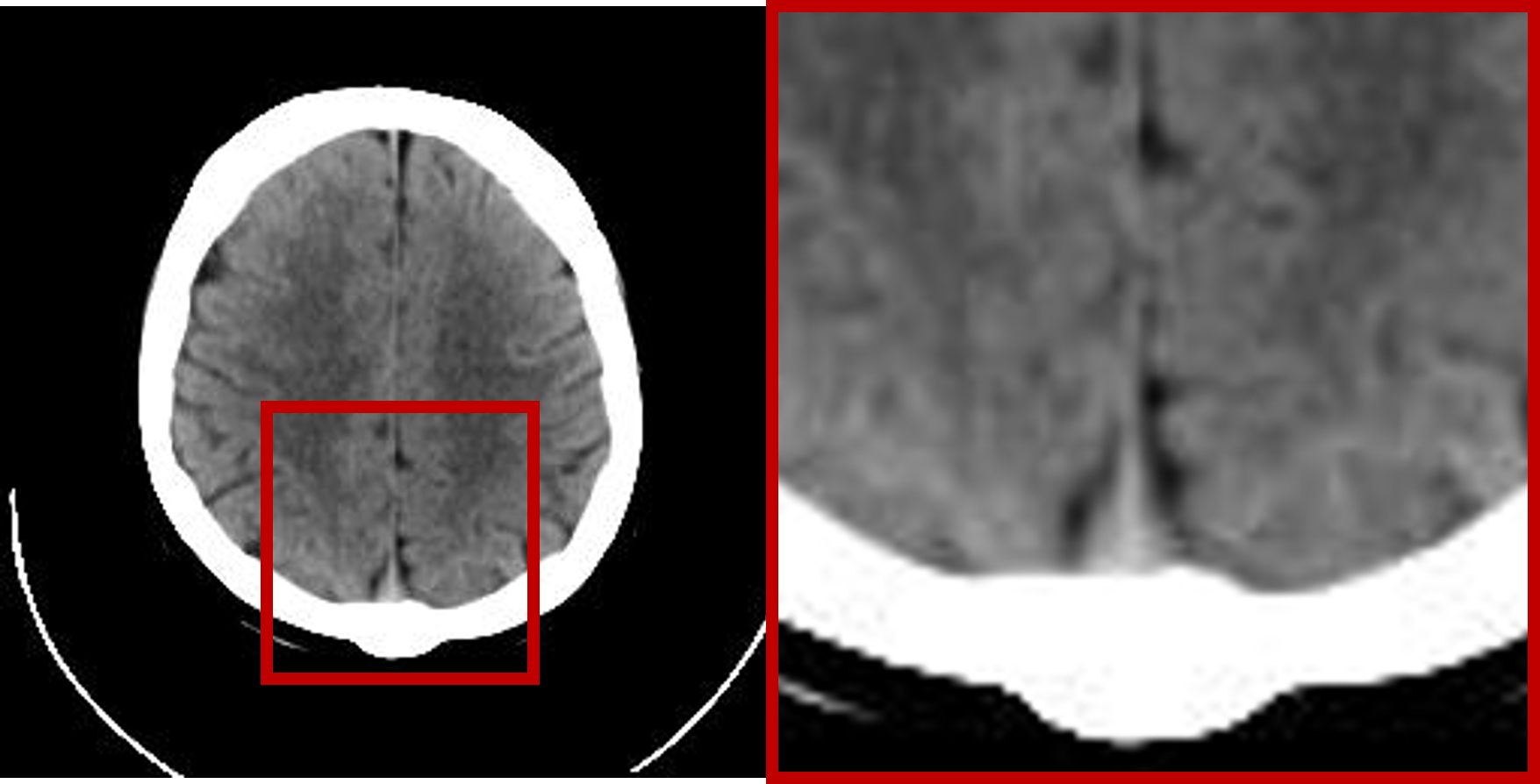}
         \caption{CT}
         \label{fig:ct_ca_orig}
     \end{subfigure}
     \hfill
          \begin{subfigure}[b]{0.24\textwidth}
         \centering
         \includegraphics[width=\textwidth]{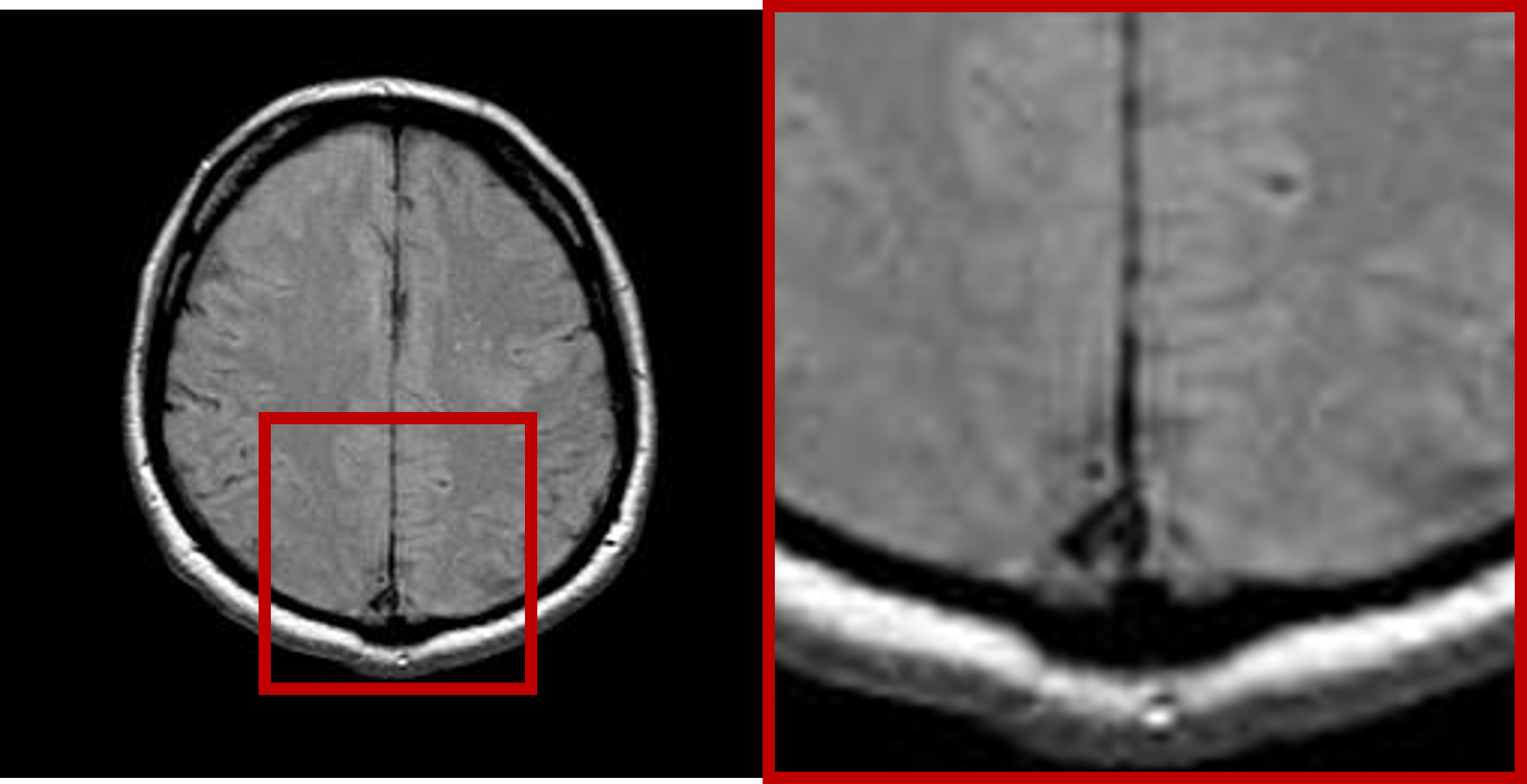}
         \caption{MRI}
         \label{fig:mri_ca_rig}
     \end{subfigure}
     \hfill
     \begin{subfigure}[b]{0.24\textwidth}
         \centering
         \includegraphics[width=\textwidth]{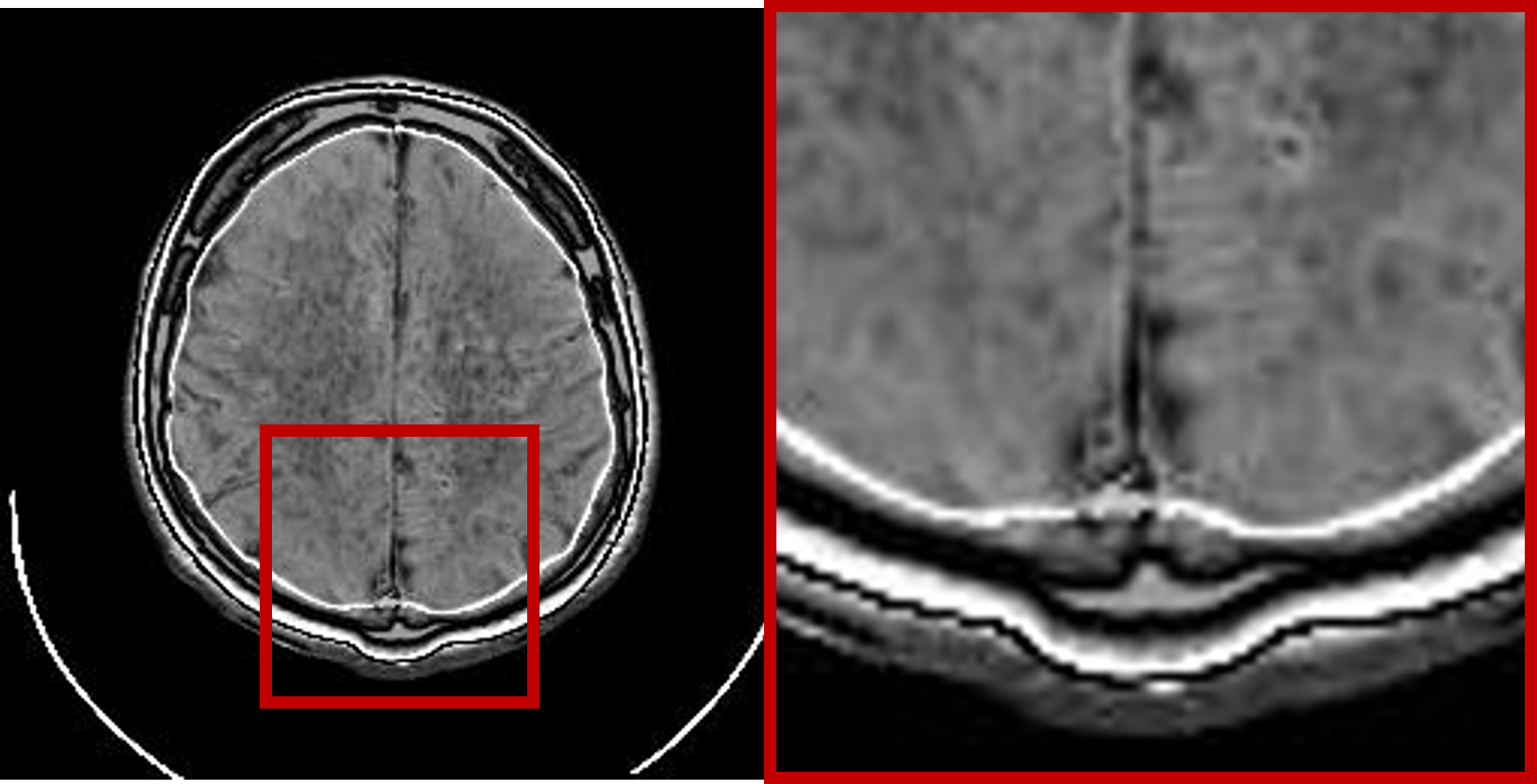}
         \caption{Zero-learning \cite{lahoud2019zero}}
         \label{fig:ca_zs}
     \end{subfigure}
     \hfill
     \begin{subfigure}[b]{0.24\textwidth}
         \centering
         \includegraphics[width=\textwidth]{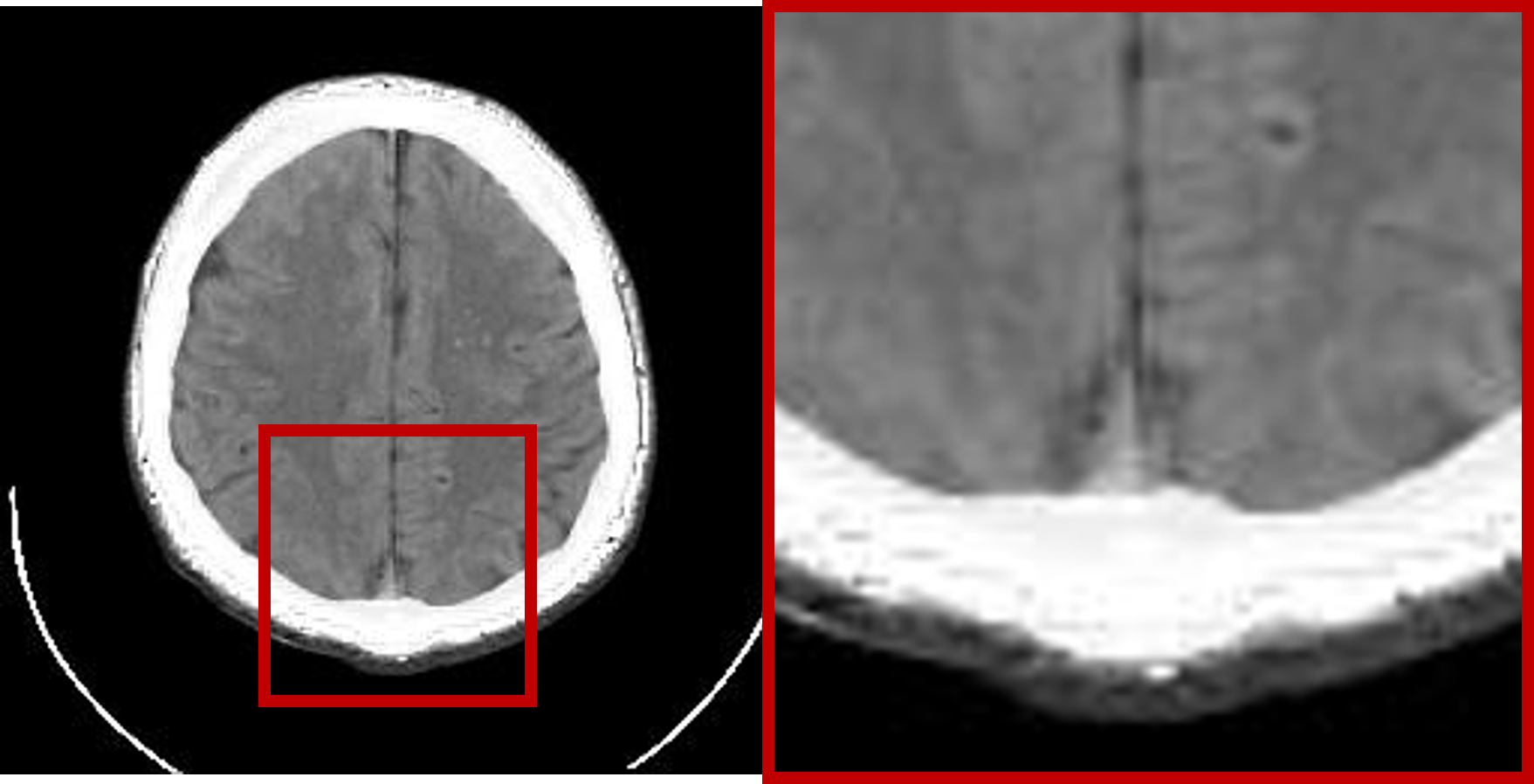}
         \caption{MSPRAN \cite{fu2021multiscale}}
         \label{fig:ca_fer}
     \end{subfigure}
     \hfill
     \begin{subfigure}[b]{0.24\textwidth}
         \centering
         \includegraphics[width=\textwidth]{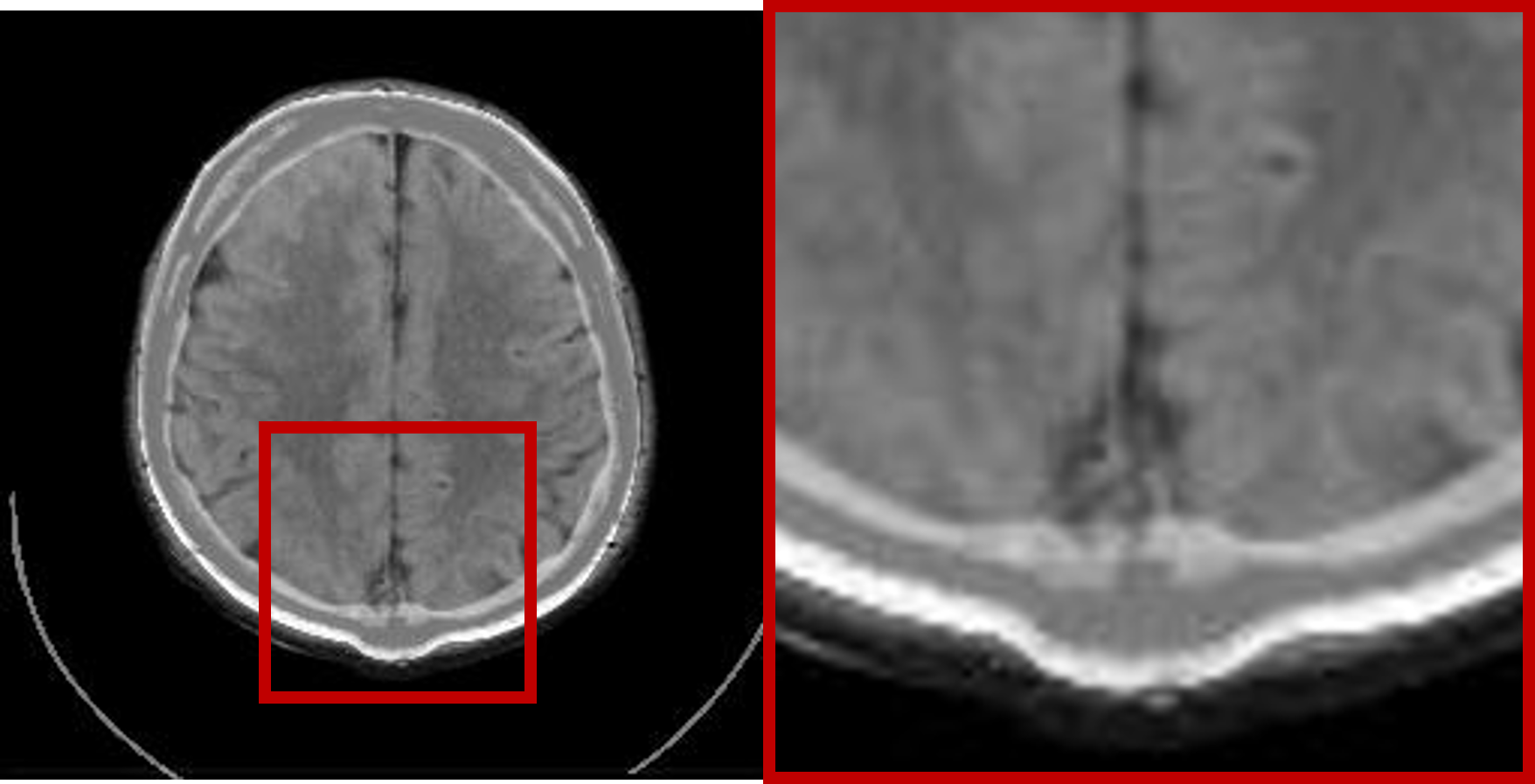}
         \caption{MSDRA \cite{li2022multiscale}}
         \label{fig:ca_fl1n}
     \end{subfigure}
     \hfill
     \begin{subfigure}[b]{0.24\textwidth}
         \centering
         \includegraphics[width=\textwidth]{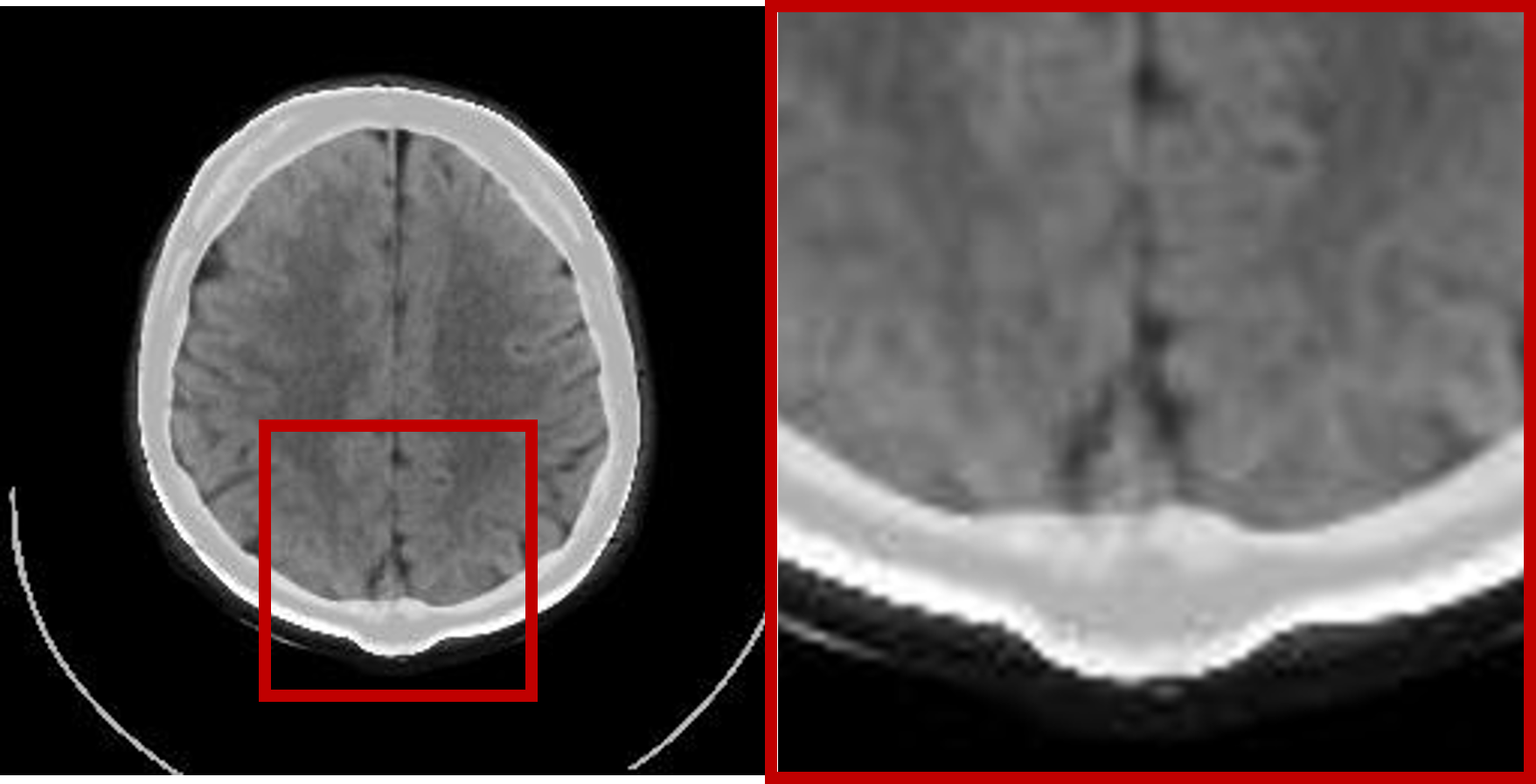}
         \caption{Ours}
         \label{fig:ca_our}
     \end{subfigure}
     \caption{Qualitative comparison of different methods between CT and MRI. (a) and (b) are source images, (c) is the zero-learning method \cite{lahoud2019zero}, (d) is the MSPRAN method \cite{fu2021multiscale}, (e) is the MSDRA method \cite{li2022multiscale}, and (f) is the method we proposed in this work.}
     \label{ca_qualres}
\end{figure}

From Figure \ref{ca_qualres}, it is also obvious from the enlarged red box that the proposed method preserves the edge and detail information well. Compared to other methods, Figure \ref{fig:ca_zs} does not retain the edge information from CT, and details of inner tissues are also distorted; Figure \ref{fig:ca_fer} has a favorable visual appearance, but the fidelity is low and lost the tissue boundary information from the MRI source image. Figure \ref{fig:ca_fl1n} has an uncleared boundary so it is hard to differentiate between boundaries and tissues. Finally, our proposed method has a better intensity contrast between edges and tissues, retains important information from both source images, and results in better fidelity.

\subsection{Ablation study}

To test the effectiveness of our proposed loss function, we ablate the image gradient loss and the perceptual loss as described in Equation \eqref{grad_img} and \eqref{perp_loss}. Hence, the loss function becomes only the MSE loss in Equation \eqref{mse}. Table \ref{abl} shows the quantitative results for the two loss functions on six different fusion metrics.

\begin{table}[h]
\renewcommand{\arraystretch}{1.3}
\caption{Ablation study on the loss function, bold numbers represent optimal values.}
\centering
\resizebox{\columnwidth}{!}{
\begin{tabular}{ccccccc}
\hline & PSNR & SSIM & MI & FMI-Pixel & FSIM & Entropy \\
\hline Only MSE & $\textbf{16.519}$ & $0.739$ & $4.428$ & $0.890$ & $\textbf{0.820}$ & $8.72$ \\
All loss & $16.413$ & $\textbf{0.740}$ & $\textbf{4.558}$ & $\textbf{0.891}$ & $\textbf{0.820}$ & $\textbf{9.816}$ \\
\hline
\end{tabular}
}
\label{abl}
\end{table}

We observe that using the ablated loss function (MSE loss) only achieves a higher PSNR value by a small margin. The rest of the metrics except FSIM, are all worse than using all three losses (Equation \eqref{total_loss}). FSIM values are equal across both loss functions. It is evident that gradient and perceptual loss will force the network to learn the high-level semantic features and focus more details on the features that would potentially contain important information, rather than purely minimizing pixel intensity differences using the traditional MSE loss. 

\section{Discussion and Limitations}

Through the comparison of different fusion results visually, we can see that the fused image using our proposed framework have a clearer representation of boundaries and details of inner tissues. The image produced by our framework is more natural than other baseline methods, where other images are either a little bit sharp, contain artificial noise, or the boundary information is not clear enough. The quantitative results validate our findings by performing well on several fusion metrics.

We do not test our proposed framework on the other two fusion tasks (MRI-SPECT and MRI-PET). The SPECT and PET images are 3-channel and MRI images are 1-channel gray-scale. To achieve the fusion task, we could transfer the 3-channel images into YCbCr color space and only focus on the Y channel for structural details and brightness information. The Y channel and the single channel MRI can be used as inputs to our framework. Then, the output can be interpreted as the fused Y-channel. We combine it with Cb and Cr channels and convert it back to RGB to get the final fused 3-channel image. We expect our proposed method also performs well on the 3-channel image fusion task. 

\begin{figure}[h!]
     \centering
          \begin{subfigure}[b]{0.24\textwidth}
         \centering
         \includegraphics[width=\textwidth]{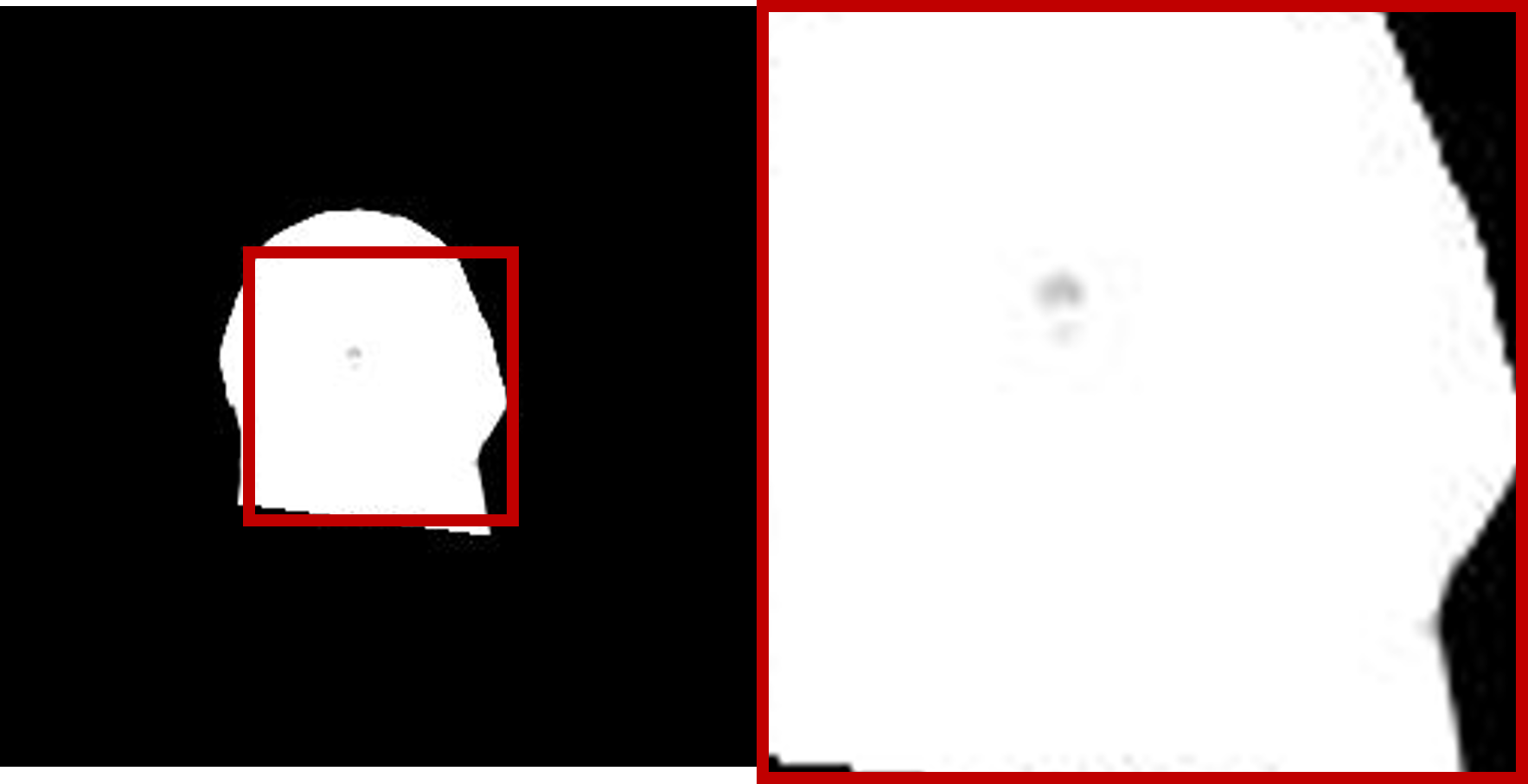}
         \caption{CT}
         \label{fig:ct_forig}
     \end{subfigure}
     \hfill
          \begin{subfigure}[b]{0.24\textwidth}
         \centering
         \includegraphics[width=\textwidth]{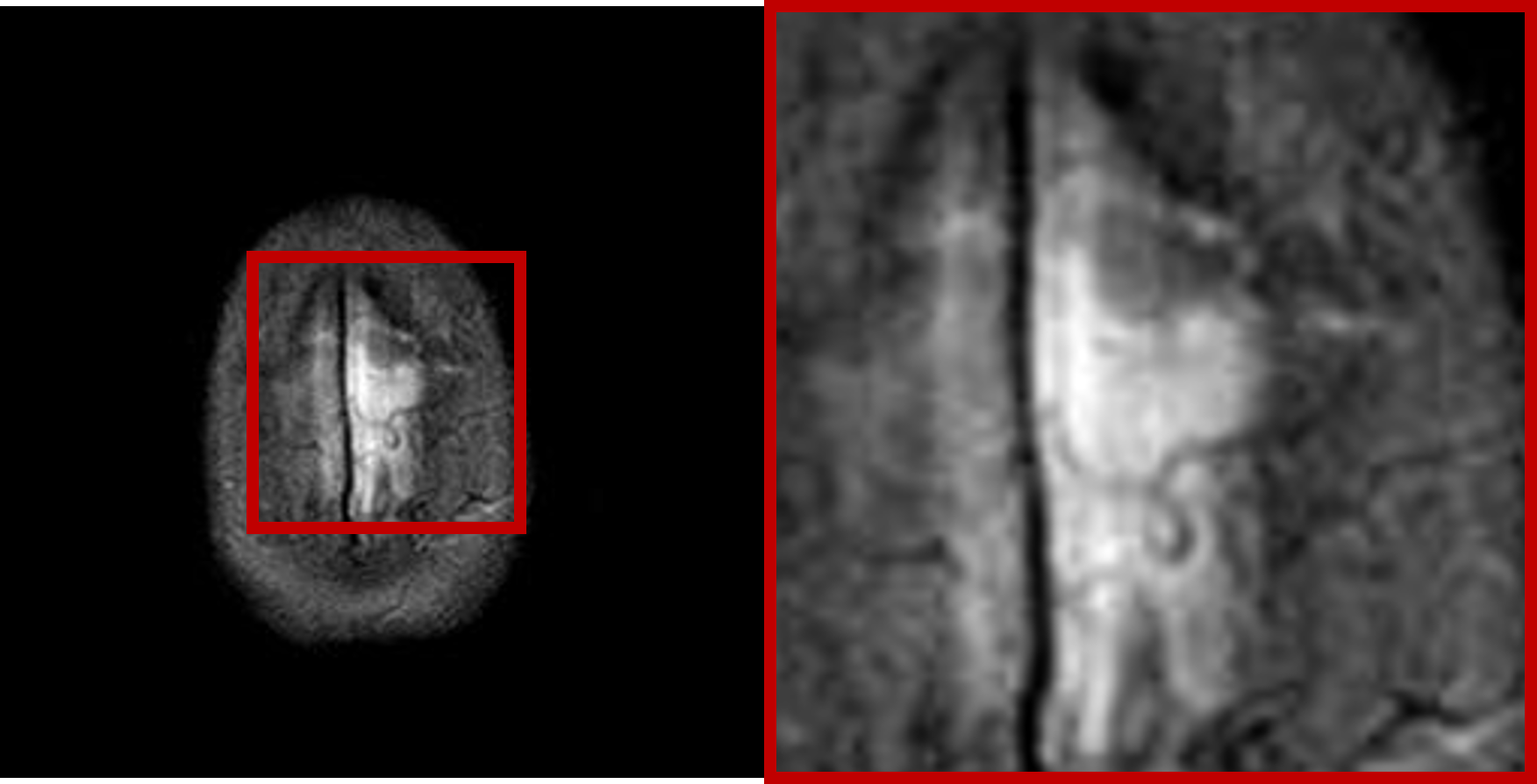}
         \caption{MRI}
         \label{fig:mri_forig}
     \end{subfigure}
     \hfill
     \begin{subfigure}[b]{0.24\textwidth}
         \centering
         \includegraphics[width=\textwidth]{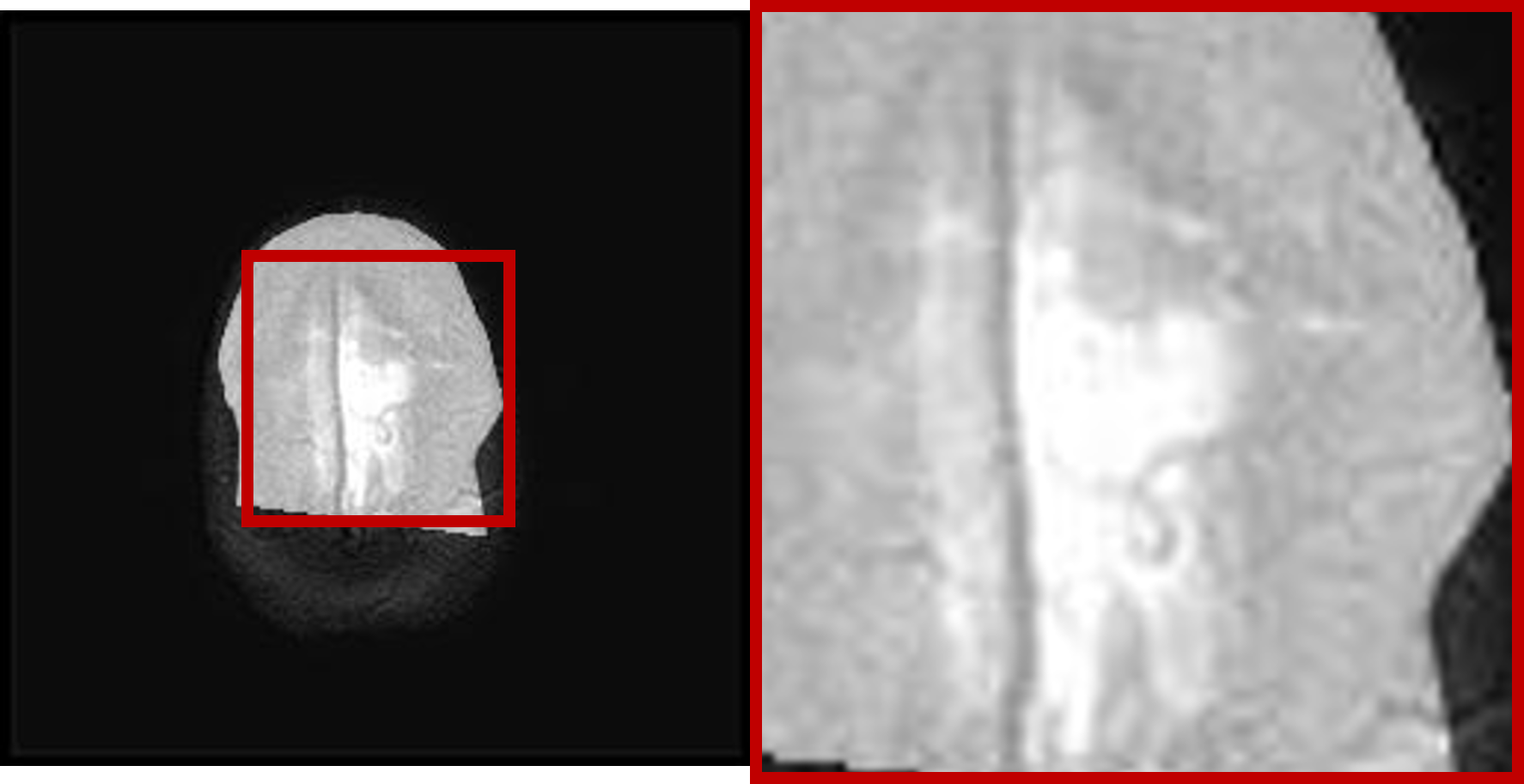}
         \caption{MSDRA \cite{li2022multiscale}}
         \label{fig:f_msdra}
     \end{subfigure}
     \hfill
     \begin{subfigure}[b]{0.24\textwidth}
         \centering
         \includegraphics[width=\textwidth]{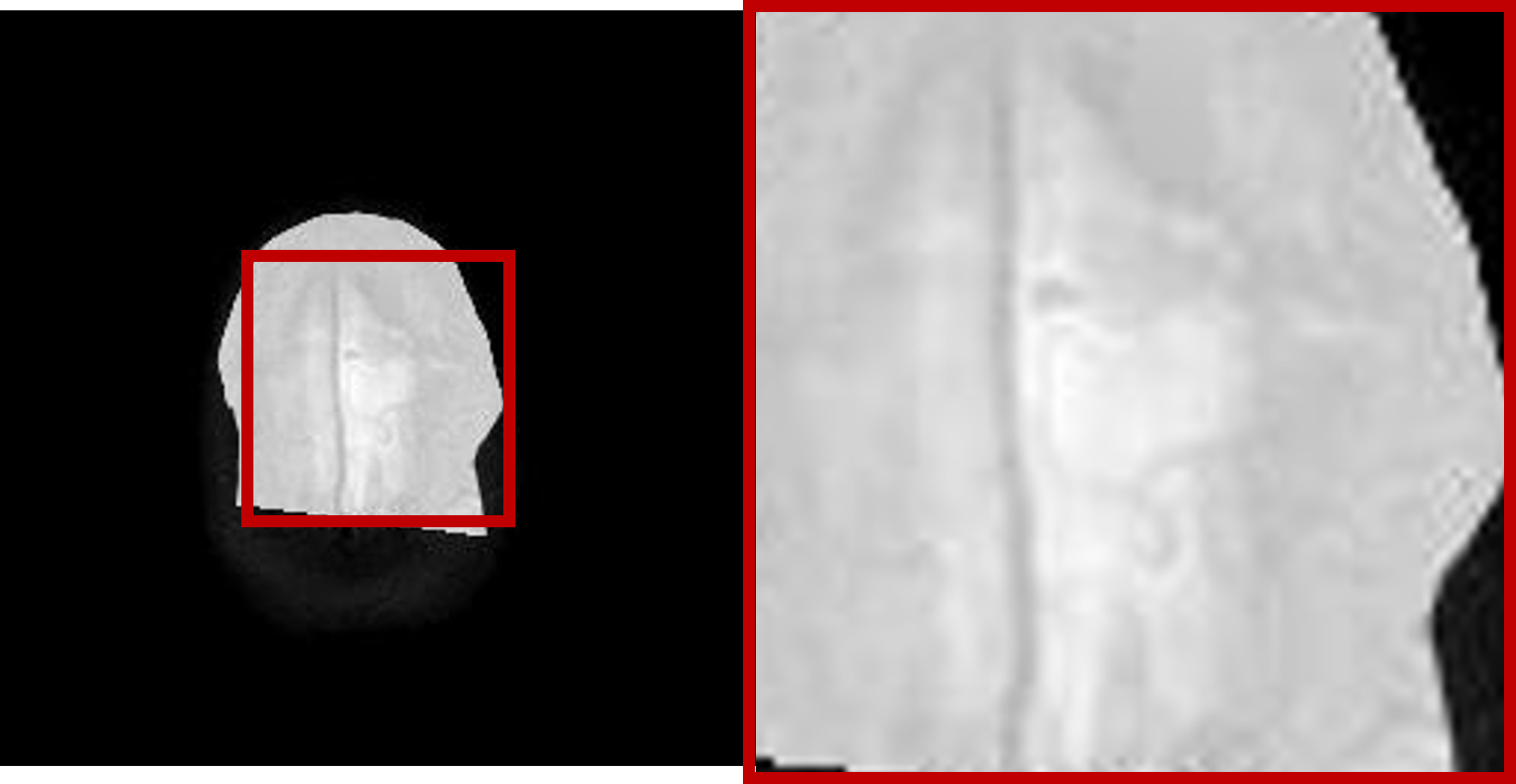}
         \caption{Ours}
         \label{fig:fours}
     \end{subfigure}
     \caption{Failure case of our proposed method, the features from CT do not handle properly and the details from MRI can barely be distinguished. MSDRA \cite{li2022multiscale} produces a slightly better visual appearance but it suffers from the artificial noise in the image.}
     \label{failed_qualres}
\end{figure}

Our proposed fusion strategy does not perform well in some cases (see Figure \ref{failed_qualres}) and may introduce fusion noise to the result. In future work, we will focus on reducing the noise while maintaining the overall fusion quality.

\section{Conclusion}

In this paper, we propose a novel network architecture based on multi-scale feature extraction and fusion strategy for multimodal medical image fusion. The original multimodal images are passed into the feature extractor to multi-scale deep semantic features. The feature maps obtained from the feature extractor are fused based on the fixed Softmax-based weighted feature fusion strategy we proposed in this work. The fused feature map is used as the input to the reconstruction module to obtain the final fused image. Our fusion strategy is fixed, which means that there is no parameter that needs to be updated in both the training and inference phase. Thus, it can achieve real-time image fusion. Extensive experiments show our proposed method is superior to several baseline methods in both subjective visual appearance and objective fusion metrics.

Our network architecture is based on the MSRPAN \cite{fu2021multiscale}, which improves on their pyramid structure. Experiments also validate the effectiveness of our method compared to other reference methods. However, our method performs not well on some corner cases and might introduce artificial noise in the fused image as we discussed previously. Despite the limitations, our method could provide a reliable reference for disease diagnosis in the real-life clinical routine.

\ifpeerreview \else
\section*{Acknowledgments}

The authors would like to thank Dr. David B. Lindell for his advice in this work and Jingtao Zhou for his suggestions on the figures.

\fi

\bibliographystyle{IEEEtran}
\bibliography{dilran_proj}








\end{document}